\documentclass[10pt,twocolumn,letterpaper]{article}

\usepackage[pagenumbers]{cvpr} 

\usepackage{tabularx}
\usepackage{booktabs}
\usepackage{array} 
\usepackage{xcolor}
\usepackage{colortbl}
\usepackage{amssymb}
\usepackage{xcolor}
\usepackage{graphicx}
\usepackage{caption}
\usepackage{makecell}
\usepackage{textcomp}
\usepackage{multirow}
\usepackage{placeins}
\usepackage{siunitx} 
\usepackage{soul}
\usepackage[most]{tcolorbox} 
\usepackage[export]{adjustbox}

\definecolor{cvprblue}{rgb}{0.21,0.49,0.74}
\definecolor{magenta}{rgb}{0.8,0,0.8}

\usepackage[pagebackref,breaklinks,colorlinks,allcolors=cvprblue]{hyperref}

\def\paperID{} 
\def\confName{CVPR}
\def\confYear{2026}

\title{VenusBench-GD: A Comprehensive Multi-Platform GUI Benchmark for Diverse Grounding Tasks}

\author{
Beitong Zhou$^{1,*}$, Zhexiao Huang$^{1,*}$, Yuan Guo$^{1,*}$, Zhangxuan Gu$^{1,*}$, Tianyu Xia$^1$, Zichen Luo$^1$, Fei Tang$^1$ \\
Dehan Kong$^2$, Yanyi Shang$^2$, Suling Ou$^1$, Zhenlin Guo$^1$, Changhua Meng$^1$, Shuheng Shen$^{1,\dagger}$ \\
$^1$Venus Team, AntGroup \quad $^2$iMean AI \\
\textcolor{magenta}{\nolinkurl{https://ui-venus.github.io/VenusBench-GD}}
}

\begin{document}
\maketitle

\footnotetext[1]{Equal contribution.}
\footnotetext[2]{Corresponding author (shuheng.ssh@antgroup.com).}

\begin{abstract}
GUI grounding is a critical component in building capable GUI agents. However, existing grounding benchmarks suffer from significant limitations: they either provide insufficient data volume and narrow domain coverage, or focus excessively on a single platform and require highly specialized domain knowledge.
In this work, we present VenusBench-GD, a comprehensive, bilingual benchmark for GUI grounding that spans multiple platforms, enabling hierarchical evaluation for real-word applications.
VenusBench-GD contributes as follows: 
\textbf{(i)} we introduce a large-scale, cross-platform benchmark with extensive coverage of applications, diverse UI elements, and rich annotated data,
\textbf{(ii)} we establish a high-quality data construction pipeline for grounding tasks, achieving higher annotation accuracy than existing benchmarks, and
\textbf{(iii)} we extend the scope of element grounding by proposing a hierarchical task taxonomy that divides grounding into \textbf{basic} and \textbf{advanced} categories, encompassing six distinct subtasks designed to evaluate models from complementary perspectives.
Our experimental findings reveal critical insights: general-purpose multimodal models now match or even surpass specialized GUI models on basic grounding tasks. In contrast, advanced tasks, still favor GUI-specialized models, though they exhibit significant overfitting and poor robustness. These results underscore the necessity of comprehensive, multi-tiered evaluation frameworks.
\end{abstract}    
\section{Introduction}
\label{sec:intro}

The rapid evolution of Multimodal Large Language Models (MLLMs)~\cite{Hurst2024GPT4oSC,claude4systemcard,qwen3technicalreport,wang2025internvl3_5} in Graphical User Interface (GUI) has necessitated robust and comprehensive benchmarks to evaluate their agentic capabilities. As a foundational and critical capability for GUI agents, GUI grounding refers to the task of aligning natural language instructions with corresponding user interface (UI) elements in a GUI screenshot or layout. The accurate grounding is a prerequisite for subsequent actions such as clicking, typing, or navigation and has drawn significant attention from researchers. As diverse efforts are ongoing in proposing innovative GUI grounding models~\cite{Gu2025UIVenusTR,Yang2025GTA1GT,hai2025holo15modelfamily,tang2025guig2,Chen2025V2PFB,He2025ReconActAS}, there are also a bunch of grounding benchmarks~\cite{Wu2024OSATLASScreenspotv2, Li2025ScreenSpotProGG, Xie2025jediosworld, Nayak2025UIVisionAD,wang2024ant,gu2023mobile,Pan2024WebCanvasBW,Wang2025MMBenchGUIHM} to evaluate their effectiveness in real-world applications.

\begin{figure}[t]
	\begin{center}
		\includegraphics[width=0.9\linewidth]{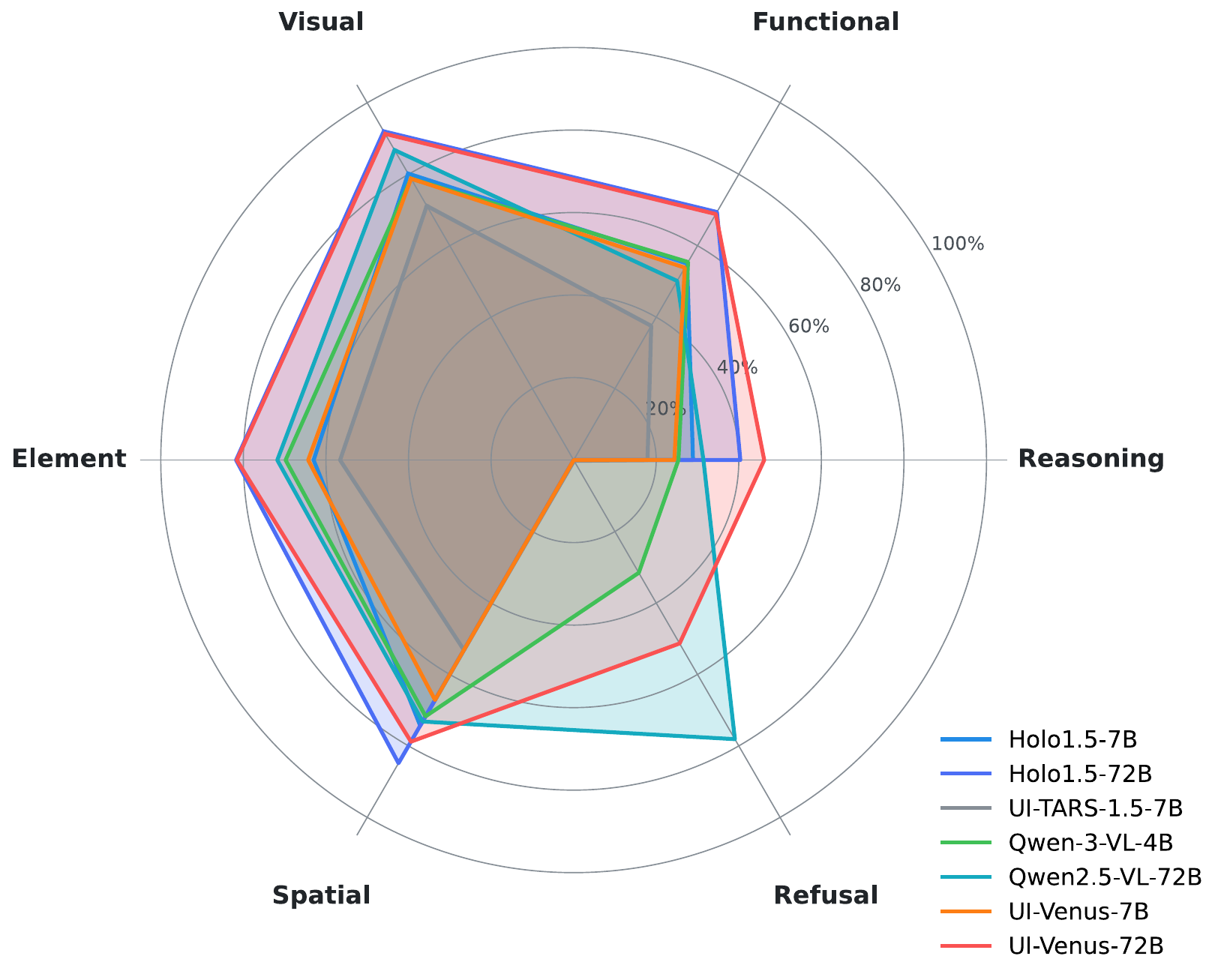}
	\end{center}
	\vspace{-5mm}
    \caption{The \textbf{mode performance} of representative GUI grounding models. Notably, model performance on advanced grounding tasks are significantly lower than on basic tasks, highlighting the increased difficulty and reasoning demands of the former.}
	\label{fig:model_radar}
	\vspace{-5mm}
\end{figure}


\begin{figure*}[!htbp]
    \begin{center}
        \includegraphics[width=1.0\linewidth]{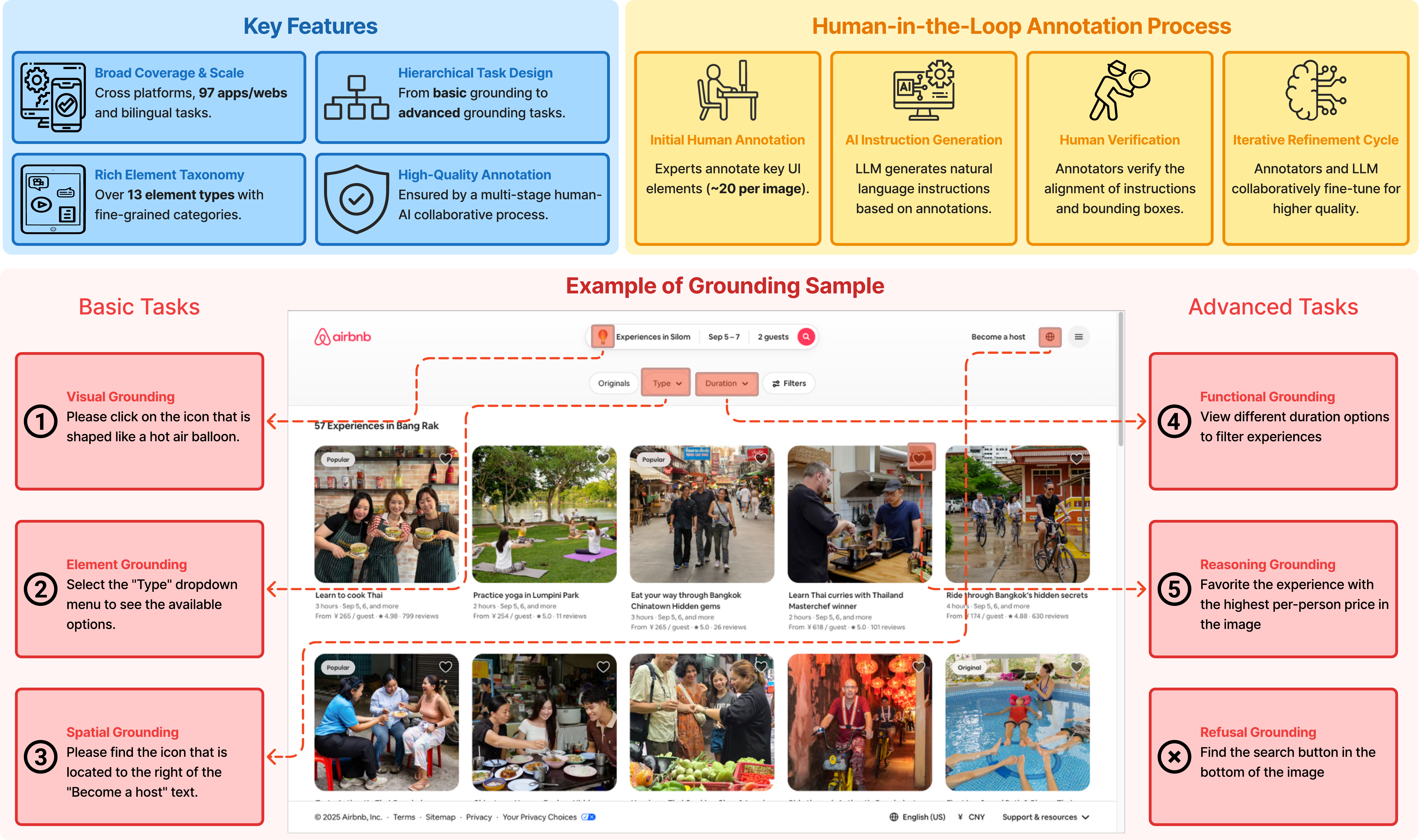}
    \end{center}
	\vspace{-5mm}
    \caption{\textbf{The overview of VenusBench-GD benchmark.} VenusBench-GD integrates basic and advanced grounding tasks to comprehensively evaluation the capabilities of existing GUI models as shown above. Basic tasks assess the ability to recognize local UI elements, while advanced tasks require holistic reasoning over the entire interface and its underlying application functionality, demanding a more complex and global understanding.}
    \vspace{-2mm}
    \label{fig:framework}
\end{figure*}

However, we argue that existing GUI grounding benchmarks suffer from several fundamental limitations that hinder their effectiveness in advancing real-world GUI agents.
First, early benchmarks like ScreenSpot-V2~\cite{Wu2024OSATLASScreenspotv2} are limited in scale and diversity of annotated UI elements, leading to nearly \textbf{95\%} accuracy by current methods~\cite{Gu2025UIVenusTR,Yang2025GTA1GT}. This results in performance saturation, making it hard to distinguish truly capable models from those overfitting to specific distributions.
Second, recent benchmarks~\cite{Li2025ScreenSpotProGG,Xie2025jediosworld,Nayak2025UIVisionAD} introduce specialized tasks requiring domain-specific knowledge (e.g., precise developer terminology or usage of professional applications), which, though technically challenging, lack ecological validity and fail to reflect the generalization needed for everyday user tasks.
What's more, narrow and simplistic grounding tasks are overwhelmingly emphasized, primarily limited to locating icons or text elements given different descriptions. In contrast, real-world GUI agent workflows involve far richer and more diverse grounding interactions. The absence of such complex and compositional tasks leads to current evaluations that capture only a fraction of the capabilities required for robust GUI understanding and action.

\begin{figure}[t]
	\begin{center}
		\includegraphics[width=0.8\linewidth]{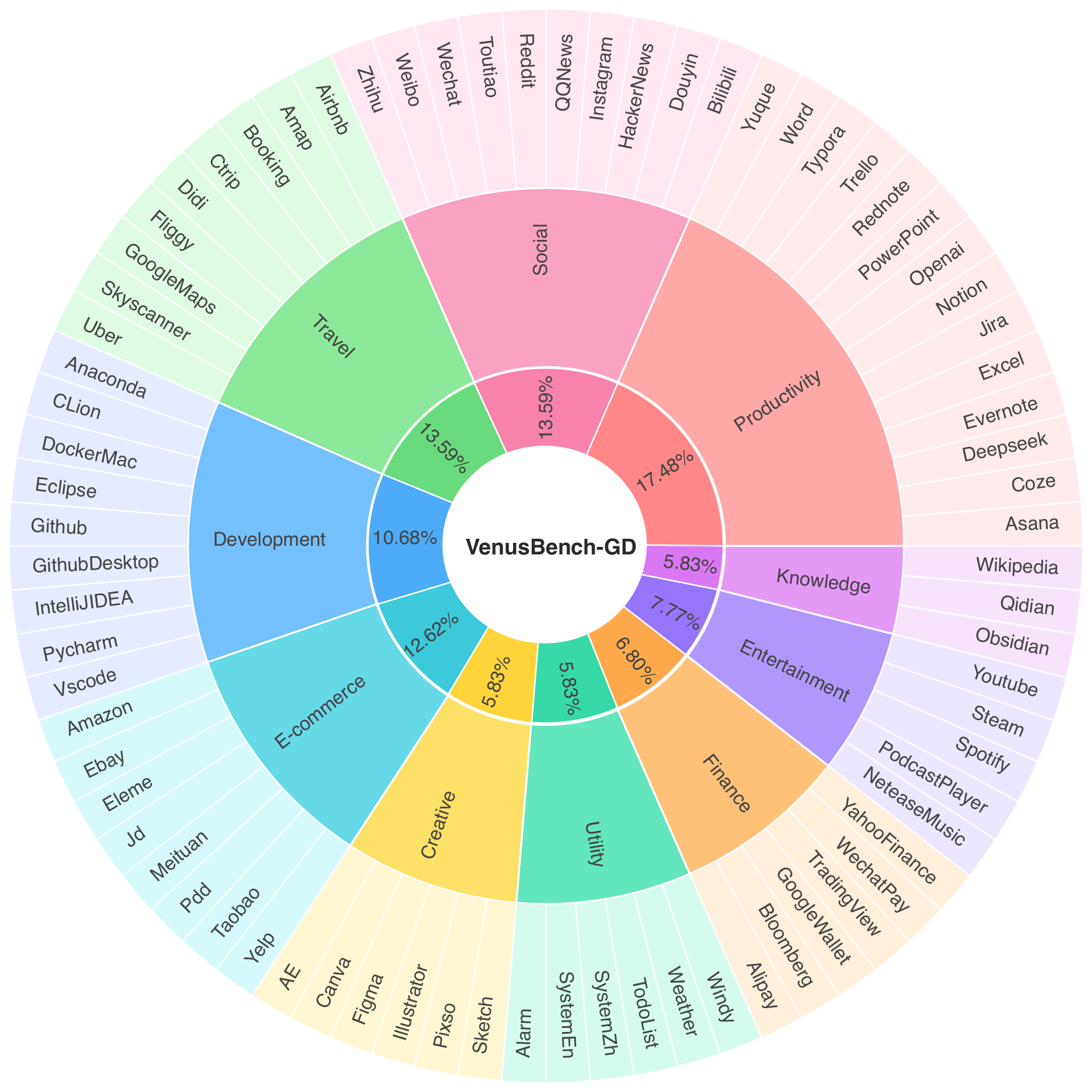}
	\end{center}
	\vspace{-5mm}
    \caption{The \textbf{domain distribution} of our grounding benchmark. VenusBench-GD spans \textbf{97} distinct apps, software, and websites across desktop, mobile, and web platforms, ensuring diverse and comprehensive coverage. We consolidate representations of the same software across platforms into one entry for clarity.}
	\label{fig:category_sunburst}
	\vspace{-5mm}
\end{figure}

To address these pressing challenges, we introduce VenusBench-GD, a large-scale, high-quality, and multi-dimensional benchmark designed to rigorously evaluate the next generation of GUI Agents. VenusBench-GD spans three major platforms: web, mobile and desktop, covering 97 distinct applications, websites and software across 10 diverse domains as shown in~\cref{fig:category_sunburst}. The benchmark comprises $6,166$ human-annotated image-instruction pairs, encompassing 13 fine-grained UI element types-including texts, buttons, sliders, tabs, links, and more-making it the largest and most semantically rich element grounding benchmark to date in terms of both scale and UI element diversity. Crucially, the instructions and screenshots are provided in both English and Chinese, enabling evaluation under multilingual and cross-cultural interaction scenarios to reflect real-world usage. To ensure fidelity, every sample in VenusBench-GD undergoes multiple rounds of verification and localization calibration, combining expert human review with model-assisted consistency checking, resulting in higher label quality compared to existing benchmarks according to our blind quantitative analysis.

Most importantly, VenusBench-GD introduces a hierarchical evaluation framework consisting of 6 distinct grounding tasks designed to probe models from complementary perspectives. Basic grounding tasks assess the fundamental perception capabilities of UI elements in the screenshot, including Element Grounding, Spatial Grounding and Visual Grounding. Advanced grounding tasks evaluate the higher-order reasoning and the robustness of models in complex scenarios, including Functional Grounding, Reasoning Grounding and Refusal Grounding. This multi-faceted design enables VenusBench-GD to evaluate not only localization accuracy but also semantic comprehension and robustness to ambiguity. In conclusion, VenusBench-GD offers a more generalizable, realistic, and challenging benchmark for next-generation multimodal grounding models.

Our experimental results reveals two key findings: 1) general MLLMs now match or even surpass GUI-specialized models on basic grounding tasks, with Qwen3-VL-8B achieving the highest average accuracy of 76.96\% among models of comparable scale; 2) Advanced grounding, particularly Reasoning and Functional Grounding, which heavily rely on domain-specific GUI knowledge, still present a clear advantage for specialized models. However, most current GUI models suffer from overfitting, as evidenced by their poor performance on Refusal Grounding.

Our major contributions are threefold:
\begin{itemize}
    \item We propose VenusBench-GD, the largest and most diverse GUI element grounding benchmark to date, spanning three platforms, 97 real-world applications across 10 domains and 13 UI element types, enabling robust cross-platform and multilingual grounding evaluation.
    \item We establish a high-quality annotation pipeline combining human review and model-assisted consistency checks, reducing label noise and improving data reliability over existing benchmarks.
    \item We propose a novel hierarchical evaluation framework comprising six complementary grounding tasks and comprehensive experimental results indicate that model performance on basic grounding tasks has nearly saturated, necessitating the introduction of more complex tasks to provide new directions for the advancement of GUI grounding models.
\end{itemize}

\begin{table*}[t]
\centering
\footnotesize
\setlength{\tabcolsep}{3pt}
\begin{tabular*}{\textwidth}{@{\extracolsep{\fill}}l *{9}{c}}
\toprule
 & \multicolumn{2}{c}{\textbf{Environments}} & \multicolumn{4}{c}{\textbf{Quality Analysis$\downarrow$}} & \multicolumn{3}{c}{\textbf{Statistics$\uparrow$}} \\
\cmidrule(lr){2-3} \cmidrule(lr){4-7} \cmidrule(lr){8-10}
\textbf{Benchmarks} & \textbf{Platform} & \textbf{Language} &
\textbf{IL} & \textbf{AR} & \textbf{SM} &\textbf{Overall} &
\textbf{\# Sample} & \textbf{Avg. Ele} & \textbf{Apps} \\
\midrule
ScreenSpot-V2~\cite{Wu2024OSATLASScreenspotv2} & Desktop, Web, Mobile & EN & 9.2\% & 0.8\% & 0.8\% &10.8\% & 1,272 & 1.74 & N/A \\
ScreenSpot-Pro~\cite{Li2025ScreenSpotProGG} & Desktop & EN,CN & 2.2\% & 1.8\% & 0.6\% &4.6\% & 1,581 & 1.0 & 23 \\
OSWorld-G~\cite{Xie2025jediosworld} & Desktop & EN & 9.2\% & 2.6\% & 2.4\% &14.2\% & 564 & 2.25 & 8 \\
UI-Vision~\cite{Nayak2025UIVisionAD} & Desktop & EN & 6.0\% & 2.4\% & 2.0\% &10.4\% & 5,479* & \textbf{4.64} & 83 \\
CAGUI~\cite{Zhang2025AgentCPMGUIBM} & Mobile & CN & 11.6\% & 6.4\% & 6.6\% &24.6\% & 3,000 & 1.0 & N/A \\
\midrule
VenusBench-GD (ours) & Desktop, Web, Mobile & EN, CN & \textbf{1.0\%} & \textbf{0.8\%} & \textbf{0.8\%} &\textbf{2.6\%} & \textbf{6,166} & 2.83 & \textbf{97} \\
\bottomrule
\end{tabular*}
\vspace{-0.2em}
\caption{Comparison of existing GUI grounding benchmarks.
Note that we only count samples related to element grounding tasks for consistency(* means UI-Vision has 8227 queries but only 5479 are about element grounding), as our evaluation focuses on GUI grounding and this may result in lower counts for other benchmarks.
\vspace{-3mm}}
\label{tab:benchmark_comparison}
\end{table*}

\section{Related Work}
\label{sec:related}

\subsection{GUI Grounding}


Driven by the remarkable progress of multimodal large language models (MLLMs)~\cite{Hurst2024GPT4oSC,Qwen2.5-VL,InternVL25} in understanding and reasoning over both visual and linguistic inputs, recent years have witnessed rapid advancements in graphical user interface (GUI) agents~\cite{Wang2024MobileAgentAM,Wang2025MobileAgentESM,Agashe2024AgentSA,droidrun,UITARS1,Gu2025UIVenusTR,Hong2023CogAgentAV,Xu2024AguvisUP}. 



Effective GUI agents rely on \textbf{GUI grounding}: the ability to locate and identify UI elements based on natural language commands. This serves as the critical link between high-level planning and low-level actions, enabling precise interactions with graphical interfaces.
Early grounding models relied on supervised fine-tuning using large labeled screenshot datasets~\cite{UITARS1,uground,Xie2025jediosworld,Wu2025GUIActorCV}, while recent approaches shift to reinforcement learning frameworks—inspired by DeepSeek-R1—that leverage continuous or spatial reward signals to enhance robustness and precision~\cite{Gu2025UIVenusTR,ui-tars-15-seed, tang2025guig2, hai2025holo15modelfamily}.
\subsection{GUI Benchmarks}


Existing benchmarks suffer from several notable limitations. Some benchmarks are constrained by limited scale, offering insufficient data to support robust model evaluation. The scale of ScreenSpot-V2~\cite{Wu2024OSATLASScreenspotv2}, while a valuable early effort in GUI grounding, is insufficient to support rigorous assessment of current models with only around $1,200$ annotated samples. 
Others are narrowly scoped to a single platform or domain-specific applications, which limits their applicability to broader user interactive scenarios. ScreenSpot-Pro~\cite{Li2025ScreenSpotProGG} targets professional desktop applications where the cluttered and tool-rich screenshots make grounding hard. UI-Vision~\cite{Nayak2025UIVisionAD} offers high-quality dataset in professional software understanding, spatial reasoning and complex agent actions, captured by trained annotators using proprietary tools. OSWorld-G~\cite{Xie2025jediosworld} samples from agent rollouts in OSWorld~\cite{Xie2024OSWorldBM} and encompasses a comprehensive set of challenges, including layout understanding, fine-grained manipulation, and infeasibility handling, among others. CAGUI~\cite{Zhang2025AgentCPMGUIBM} is a Chinese Android GUI benchmark for both grounding and agent tasks enabling multilingual evaluation beyond English-only datasets. 

Among existing benchmarks, MMBench-GUI~\cite{Wang2025MMBenchGUIHM} is the closet in scope to our work. However, its element grounding subset comprises only around $3,500$ samples and adopts a coarse task taxonomy that merely divides the instructions into ``basic" and ``advanced" categories. This limited granularity constraints its ability to capture fine-grained variations in task complexity. Moreover, they lack annotation quality check and error case analysis, which may affect the objectivity of benchmark validation in assessing GUI grounding capabilities. 

\newtheorem{definition}{Definition}
\section{VenusBench-GD}




\subsection{Tasks Design}\label{subsec:3.1}

GUI grounding is a fundamental capability of GUI agents~\cite{UITARS1,ui-tars-15-seed,tang2025guig2,Gu2025UIVenusTR,hai2025holo15modelfamily}. Intuitively, it refers to the task of mapping user-provided natural language instructions to specific boxes or regions within an interface image~\cite{Wu2024OSATLASScreenspotv2,Li2025ScreenSpotProGG}. 
As a natural extension, the grounding function can also be defined to output a point rather than a region. 

Inspired by human interaction with digital interfaces, where users jointly rely on global context to grasp app functionality and local perception to identify target UI elements, we design element grounding tasks in VenusBench-GD with \textbf{basic grounding} focusing on local understanding (\emph{e.g.}, element type, text, appearance, and local spatial relations) and \textbf{advanced grounding} requiring global contextual reasoning and functional comprehension of the application.

While prior work~\cite{Wu2024OSATLASScreenspotv2,Li2025ScreenSpotProGG} has largely addressed basic element grounding tasks about icons or texts, we extend this foundation by enriching the task taxonomy, incorporating visual grounding, and formally define our \textbf{basic grounding} tasks herein.

\begin{itemize}
    \item \textbf{Element Grounding}: The model locates a UI element based directly on its type (\emph{e.g.}, button, dropdown) or associated text content via OCR. This is the most common task in prior benchmarks and primarily evaluates basic UI comprehension.
    \item \textbf{Spatial Grounding}: Building on element grounding, this task requires the model to interpret relative spatial cues, such as “left of,” “below,” or “in the same row as”, to identify the target element, thereby assessing fine-grained localization within local layout contexts.
    \item \textbf{Visual Grounding}: The model grounds elements based on visual appearance descriptions, such as shape, color, or icon pattern (\emph{e.g.}, “the icon shaped like a hot air balloon”), testing its ability to perceive visual attributes.
\end{itemize}

This tiered design allows systematic evaluation of visual grounding at different abstraction levels. However, the basic tasks focus only on foundational OCR or visual detection capabilities, missing the advanced reasoning and global screenshot understanding demonstrated by modern GUI agents.
To address this gap, we introduce three novel \textbf{advanced grounding} tasks: Reason Grounding, Functional Grounding, and Refusal Grounding, each requiring more complex reasoning abilities.

\begin{itemize}
    \item \textbf{Reasoning Grounding}: 
    The model must reason over visible information (e.g., comparing prices, calculating dates) to identify the correct UI element. Unlike direct matching, this task involves intermediate cognitive steps such as numerical or logical inference (\emph{e.g.}, “select the cheapest option”).
    \item \textbf{Functional Grounding}: 
    The target element is specified by its function (\emph{e.g.}, “close the video” or “add to cart”), requiring the model to understand common UI affordances. To ensure broad applicability, we focus on generic rather than domain-specific functionalities.
    \item \textbf{Refusal Grounding}: 
    The instruction refers to a non-existent UI element, and the model must refrain from predicting a bounding box. This task reflects a model’s resistance to overfitting and supports fairer evaluation of GUI grounding capabilities.
\end{itemize}

In conclusion, Figure~\ref{fig:framework} demonstrates the specific instructions for the types of grounding task using a concrete image example, which helps to clarify the hierarchical evaluation in VenusBench-GD. To better explain all six grounding tasks, we show the input prompt as follows while the red sentence refer to the refusal grounding.

\begin{tcolorbox}[
    title=Grounding Prompt,breakable]
\small {Output the center point of the position corresponding to the instruction: \texttt{\color{red} \{\}}. The output should just be the coordinates of a point, in the format [x,y]. \texttt{\color{red} Additionally, if you think the task is infeasible (e.g., the task is not related to the image), the output should be [-1,-1].}} 
\end{tcolorbox}

{
\setcellgapes{1.5pt}
\makegapedcells

\begin{table}[t]
\centering
\footnotesize
\setlength{\tabcolsep}{3pt}
\begin{tabular}{c c l}
\toprule
\textbf{Grounding Tasks} & \textbf{Count} & \textbf{Examples} \\
\midrule

Element Grounding & 1591 & \makecell[l]{Locate the \textbf{text} "Liu Chun". \\ Find the "Image" \textbf{menu item}.} \\
Spatial Grounding & 1029 & \makecell[l]{Find the user settings icon \\ in the \textbf{lower-left corner}.} \\
Visual Grounding & 839  & \makecell[l]{Click the \textbf{heart-shaped} icon located \\ in the bottom of the image gallery.} \\

\midrule

Reasoning Grounding & 1107 & \makecell[l]{Book a ticket with the \textbf{third-highest} \\ price and \textbf{the latest} departure time.} \\
Functional Grounding & 700  & \makecell[l]{Quickly enable the device's GPS \\ positioning function} \\
Refusal Grounding & 900  & \makecell[l]{Find the blue circular icon in the \\ upper \textbf{\st{left} right} corner.} \\

\bottomrule
\end{tabular}
\caption{Examples of each grounding task and the key features are highlighted.}
\label{tab:grounding_tasks}
\vspace{-3mm}
\end{table}
}

\begin{figure}[t]
	\begin{center}
		\includegraphics[width=0.8\linewidth]{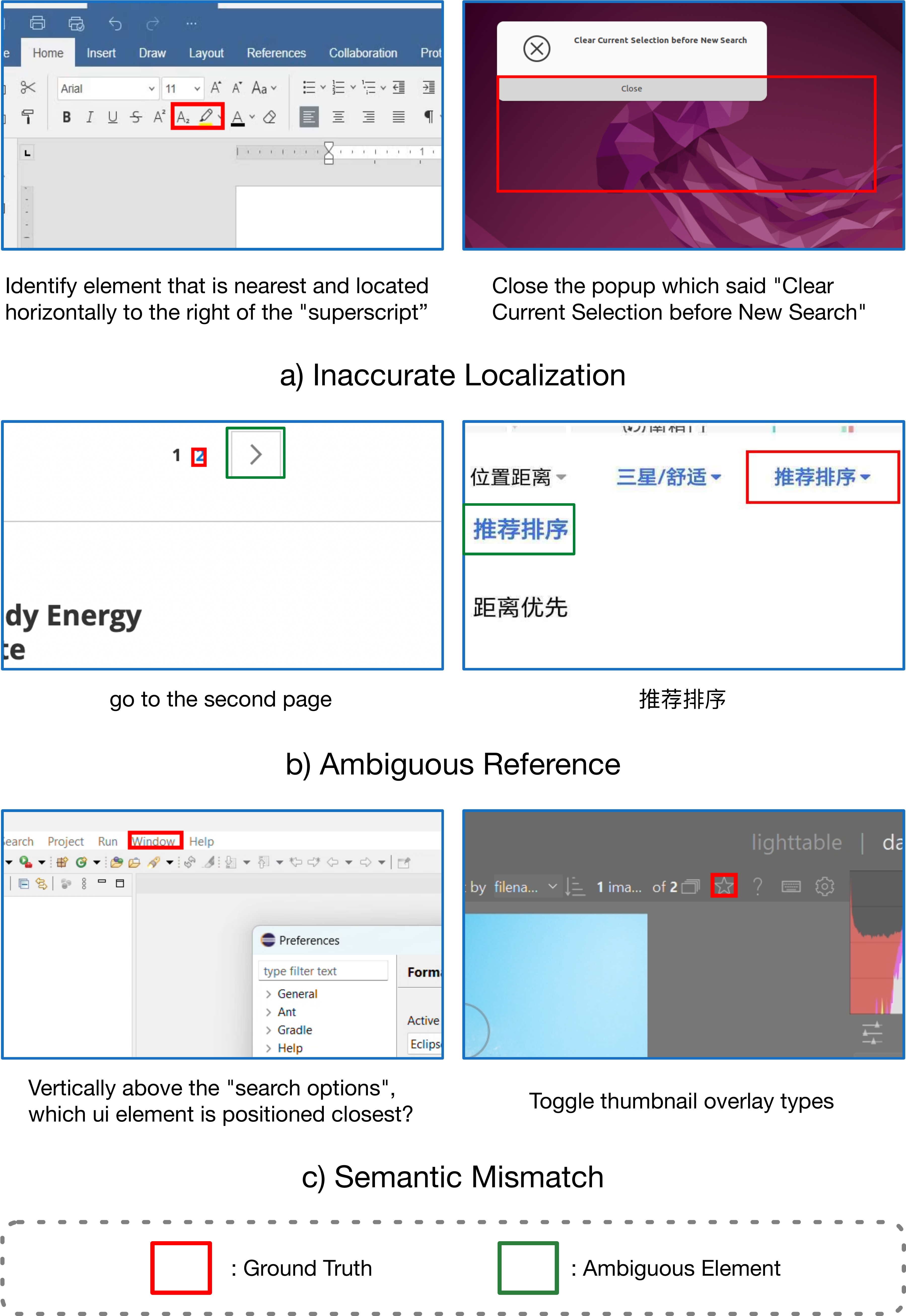}
	\end{center}
	\vspace{-5mm}
    \caption{Examples of inaccurate annotations in existing benchmarks.}
	\label{fig:quality_analysis}
	\vspace{-5mm}
\end{figure}

\subsection{Bottom-up Data Pipeline}\label{subsec:3.2}

To construct VenusBench-GD, we designed a bottom-up four-stage data creation pipeline combines automation and human expertise. We first gathered raw interface screenshot data and then used detectors to get various UI elements. Building upon these candidate elements, we implemented hierarchical instruction generation with MLLMs to produce task-aligned prompts for both basic and advanced grounding scenarios. Finally, we employed multi-stage filtering to select only the highest-quality and non-trivial samples. This intensive process ensures exceptional data richness, semantic diversity, and annotation fidelity.

\noindent \textbf{Raw Data Collection:} We first collected 97 distinct applications and websites from Chinese and English communities spanning 10 domains as shown in Figure~\ref{fig:category_sunburst}, including creative, utility, development, and others. Notably, versions of the same application across different platforms (\emph{e.g.}, Reddit on web vs. mobile) are counted as separate applications. A full list of applications is provided in the appendix.

\begin{table}[t]
\centering
\footnotesize
\setlength{\tabcolsep}{8pt}
\begin{tabular}{c c c c}
\toprule
\textbf{Platform} & \textbf{Applications} & \textbf{Language} & \textbf{Images} \\
\midrule
\multirow{2}{*}{Web}      & \multirow{2}{*}{36} & English & 240 \\
                          &                     & Chinese & 209 \\
\midrule
\multirow{2}{*}{Mobile}   & \multirow{2}{*}{40} & English & 296 \\
                          &                     & Chinese & 704 \\
\midrule
\multirow{2}{*}{Desktop}  & \multirow{2}{*}{21} & English & 466 \\
                          &                     & Chinese & 257 \\
\bottomrule
\end{tabular}
\caption{Distribution of our collected images, languages, and applications across platforms in VenusBench-GD.}
\label{tab:platform_app_lang_stats}
\vspace{-5mm}
\end{table}

\noindent \textbf{UI Element Localization:} To obtain a large and semantically diverse set of high-quality UI element annotations, we first defined 13 common UI element types and employed two complementary annotation strategies: (1) generating coarse bounding boxes using detectors like APT~\cite{gu2023mobile}, LayoutLM~\cite{xu2020layoutlm}, and OmniParserV2~\cite{yu2025omniparser}, followed by deduplication and human verification of localization accuracy; and (2) having annotators explicitly label long-tail, model-challenging elements (\emph{e.g.}, checkboxes, sliders). In both cases, annotators were instructed to tightly align the boxes with the interactive regions of the elements. Each box underwent at least two rounds of review, with the second round requiring $\geq 95\%$ inter-annotator accuracy. This intensive protocol yields at least 20 high-precision candidate element annotations per image.

\noindent \textbf{Instruction Generation:} For basic grounding tasks, we utilize Qwen2.5-VL-72B model~\cite{Qwen2.5-VL} to generate instructions based on collected UI element candidates, conditioned on element type, text, position, and visual appearance, and have annotators verify the instruction–element alignment, assigning valid pairs to their corresponding basic grounding subtasks. 

Instruction generation for advanced grounding tasks is more intricate. For Refusal Grounding, we deliberately perturb matched instruction–element pairs, modifying element type, text, spatial references, or visual descriptions, to ensure that the revised instruction has no valid referent in the screenshot, thereby maximizing ambiguity for testing model robustness. For Reasoning Grounding, we focus on application domains that commonly involve numerical comparison, date filtering, or item management—such as e-commerce, travel booking, and finance. Within these domains, we develop question templates and instruct annotators to generate instructions and corresponding ground truth elements only when a valid reasoning scenario exists. For Functional Grounding in advanced tasks, we used elements from the element candidates and provided both pre-click and after-click screenshots to Qwen2.5-VL-72B for generating potential functions or user intentions from these sequential images and convert them into textual instructions for further human verification.

\noindent \textbf{Post Selection:} Although multiple rounds of screening and verification have been applied, a large pool of candidate instructions remains. We therefore leverage state-of-the-art(SOTA) GUI grounding models and aggregate their predictions to filter out overly simple samples and further eliminate ambiguous ones.
Taking basic grounding tasks as an example, we generated approximately $28,000$ annotated UI elements across ~$2,000$ images. After model-based instruction generation and rigorous filtering, only $3,459$ high-quality samples were retained.

\subsection{Data Statistics}\label{subsec:3.3}

\noindent\textbf{Scale Statistics:} As shown in Table~\ref{tab:benchmark_comparison}, we compare existing benchmarks across different environments and scale statistics. VenusBench encompasses all major platforms and features the largest scale to date, with 6,166 annotated samples and the broadest application coverage among existing GUI grounding benchmarks.


\begin{table*}[h]
\centering
\setlength{\tabcolsep}{3pt}
\begin{tabular}{c|cc|c|cc}
\toprule
\multirow{3}{*}{\textbf{Benchmark}} & \multicolumn{2}{c|}{\textbf{Resolution}}              & \textbf{Element size}  & \multicolumn{2}{c}{\textbf{Instruction}}                                                                                                \\
                           & \#types & max / min                 & min / max       & \begin{tabular}[c]{@{}c@{}}characters\\ min / max / avg\end{tabular} & \begin{tabular}[c]{@{}c@{}}words\\ min / max / avg\end{tabular} \\
\midrule
ScreenSpot-V2              & 94                & $2880\times1800$ / $362\times313$   & 1.8$e^{-4}$ / 7.2$e^{-1}$ & 4 / 87 / 23.1                                                        & 1 / 16 / 4.1                                                    \\
ScreenSpot-Pro             & 14                & $6016\times3384$ / $1920\times1080$ & 1.7$e^{-5}$ / 4.7$e^{-2}$ & 2 / 102 / 26.2                                                       & 1 / 16 / 4.6                                                    \\
OSWorld-G                  & 3                 & $1920\times1080$ / $1280\times720$  & 5.8$e^{-6}$ / 7.9$e^{-1}$ & 3 / 112 / 36.3                                                       & 1 / 19 / 6.4                                                    \\
UI-Vision                  & 100               & $3360\times2036$ / $1092\times614$  & 0 / 3.1$e^{-2}$      & 3 / 173 / 49.6                                                       & 1 / 34 / 8.25 \\
VenusBench-GD              & 65                & $5120\times 2880$ / $320\times242$   & 2.7$e^{-5}$/ 3.5$e^{-1}$ & 4 / 218 / 57.1                                                       & 1 / 34 / 10.1                                                   \\
\bottomrule
\end{tabular}
\caption{Distributions and diversities of existing GUI grounding benchmarks viewed on Resolution, Element size and Instruction.}
\vspace{-3mm}
\label{table:diversity}
\end{table*}






\noindent\textbf{Diversity Statistics:} Another important factor in building an effective benchmark is the diversity of samples. In this section, we calculate the main metrics including maximum, minimum, and average based on three views, \emph{i.e.}, \textit{screen resolutions}, \textit{element sizes}, and \textit{instructions}, as shown in Table~\ref{table:diversity}. Statistical analysis of VenusBench-GD compared to prominent existing benchmarks reveals several critical advantages in design diversity and challenge level. First, while possessing 65 distinct screen sizes, VenusBench-GD exhibits by far the largest screen resolution range, providing a significantly more rigorous test for model robustness to input resolution scaling. Second, VenusBench-GD, alongside OSWorld-G, demonstrates the most substantial variation in the relative area (\emph{w.r.t.} the entire screen) of annotated UI elements. This characteristic inherently increases the difficulty of multi-scale element recognition. Third, VenusBench-GD features the longest average instruction length among all six evaluated benchmarks. The statistics serve as a strong proxy for the richness and semantic complexity of the grounding tasks within our benchmark, reflecting its higher linguistic diversity.

\subsection{Data Quality Analysis}\label{subsec:3.4}

\noindent \textbf{Qualitative Analysis:} Despite the quality check by annotators are made on existing GUI grounding benchmarks, some inaccurate annotations still exist according to our observation, which will harm the performance evaluation of various recent GUI Agents. To better explain the inaccurate, we divide common mismatches from recent GUI grounding benchmarks into three folds. When determining the optimal bounding box for a UI element during human verification, we annotate all plausible corresponding regions to capture its full spatial extent.

\begin{itemize}
    \item \textbf{Inaccurate Localization(IL)}: The annotated bounding box partially covers the correct UI element - either too large or too small. If the misaligned region (as visually estimated by annotators) exceeds roughly 40\% of the element's true area, the localization is considered inaccurate.
    \item \textbf{Ambiguous Reference(AR)}: The instruction could reasonably refer to multiple valid UI elements, but only one is labeled as the ground truth. This introduces ambiguity, as other equally plausible targets are ignored.
    \item \textbf{Semantic Mismatch(SM)}: The annotated box does not overlap with any UI element that matches the instruction’s semantics—indicating a complete misalignment between language and visual grounding.
\end{itemize}

Specifically, as illustrated in Figure~\ref{fig:quality_analysis} a), two Inaccurate Localization sampled cases reveal ground truth annotations that either encompass extraneous icons or exhibit systematic spatial offset. Another typical problem is Ambiguous Reference \emph{e.g.}, as two cases indicated in Figure~\ref{fig:quality_analysis} b). Finally, Figure~\ref{fig:quality_analysis} c) introduces Semantic Mismatch, where the predicted element does not semantically correspond to the intent of the instruction.

\noindent\textbf{Blind Quantitative Analysis:} Beyond qualitative insights, we also performed a quantitative statistical analysis of the benchmark qualities in Table~\ref{tab:benchmark_comparison}. To ensure a fair comparison of annotation quality across benchmarks, we randomly sampled $500$ instances from each of six closely related benchmarks (totaling $3,000$ samples), pooled them into a unified set, and assigned them to annotators—who had been trained on a small calibration subset—without revealing the source benchmark of each sample. 
Based on this test, we can make the following statements:

\begin{itemize}
    \item Inaccurate Localization constitutes the predominant source of inaccurate samples across all benchmarks, accounting for the highest proportion of mislabeled instances in each evaluated dataset.
    \item VenusBench-GD achieves the lowest error rates (\textbf{2.6\%}) across all three types of inaccurate cases, with particularly significant improvement in IL: reducing the error rate from 11.6\% (observed in CAGUI) to 1.0\%. This demonstrates the superior accuracy of our grounding benchmark and the fairness of GUI Agent evaluation.
\end{itemize}

Consistent with our findings, UI-Ins~\cite{Chen2025UIInsEG} reports that approximately 23.3\% of samples in existing grounding datasets suffer from annotation issues, mutually corroborating the reliability of our quality analysis.

\newcounter{remarkcounter}
\newtheorem{remark}[remarkcounter]{Remark}
\section{Experiments}

\subsection{Experiment Settings}
We evaluated a comprehensive set of both open-source and closed-source vision-language models (VLMs) and GUI agents with demonstrated capabilities.
We benchmark both closed-source and open-source models. Among open-source models, we focus on open-source VLMs and two categories of GUI-specific models: (1) GUI-specific Models (SFT) and (2) GUI-specific Models (RL).
For each model, we follow the recommended prompting format to ensure fair and optimal performance evaluation; the specific prompts used are detailed in the Appendix.

Consistent with previous works~\cite{Li2025ScreenSpotProGG,Xie2025jediosworld,Cheng2024SeeClickHG}, we define a prediction as \emph{correct} when the predicted point lies within the ground-truth bounding box and report the accuracy averaged over all UI elements. For refusal grounding tasks, we consider a model’s output correct only when it adheres strictly to the prescribed refusal format, such as returning coordinates $[-1, -1]$ or a designated rejection phrase as specified in the prompt.

\subsection{Benchmark Results}
\begin{table*}[ht]
    \centering
    \footnotesize
    \setlength{\tabcolsep}{0pt}
    \begin{tabular*}{\textwidth}{@{\extracolsep{\fill}}l *{9}{c}}
    \toprule
    \multirow{2}{*}{\textbf{Models}} & 
    \multicolumn{4}{c}{\textbf{Basic Tasks}} & 
    \multicolumn{4}{c}{\textbf{Advanced Tasks}} & 
    \multirow{2}{*}{\textbf{Overall}} \\
    \cmidrule(lr){2-5} \cmidrule(lr){6-9}
                    & \textbf{Element} & \textbf{Visual} & \textbf{Spatial} & \textbf{Avg} &
                      \textbf{Reasoning} & \textbf{Functional} & \textbf{Refusal} & \textbf{Avg} & \\
    \midrule
    \rowcolor{gray!15}
    \multicolumn{10}{l}{\textit{Closed-source Models}} \\
    GPT-4o~\cite{Hurst2024GPT4oSC}                           & 4.53 & 10.61 & 6.61 & 6.62 & 2.98 & 4.14 & \textbf{43.00} & 16.59 & 10.99 \\ 
    Gemini 2.5 Pro~\cite{gemini25pro}                        & \ul{8.99} & \ul{13.59} & \ul{11.47} & \ul{10.84} & \ul{6.50} & \ul{10.00} & \ul{40.67} & \textbf{18.77} & \ul{14.32} \\
    Claude-Sonnet-4.0~\cite{claude4systemcard}               & \textbf{12.13} & \textbf{37.78} & \textbf{14.58} & \textbf{19.08} & \textbf{11.11} & \textbf{11.71} & 30.33 & \ul{17.65} & \textbf{18.45} \\
    \midrule
    \rowcolor{gray!15}
    \multicolumn{10}{l}{\textit{General Open-source VLMs}} \\
    Qwen2.5-VL-7B~\cite{Qwen2.5-VL}                          & 58.14 & 61.26 & 53.06 & 57.39 & 14.72 & 32.00 & 18.11 & 20.32 & 41.12 \\ 
    Qwen2.5-VL-72B~\cite{Qwen2.5-VL}                          & 71.78 & \textbf{86.77} & 73.18 & 75.83 & \textbf{31.44} & 50.14 & \textbf{78.11} & \textbf{51.79} & \textbf{65.28} \\ 
    Qwen3-VL-4B~\cite{qwen3technicalreport}                          & 69.77 & 78.78 & 71.72 & 72.54 & 25.38 & \ul{55.43} & 31.55 & 35.20 & 56.15 \\ 
    Qwen3-VL-8B~\cite{qwen3technicalreport}                  & \ul{73.48} & 83.31 & \ul{77.16} & \ul{76.96} & 24.75 & 47.71 & \ul{63.78} & \ul{43.66} & \ul{62.34} \\ 
    Qwen3-VL-30B-A3B~\cite{qwen3technicalreport}             & \textbf{76.24} & \ul{86.53} & \textbf{77.45} & \textbf{79.10} & \ul{30.17} & \textbf{61.86} & 25.56 & 36.83 & 60.54 \\ 
    InternVL2.5-8B~\cite{InternVL25}                        & 13.76 & 18.24 & 14.48 & 15.06 & 7.05 & 11.71 & 0.33 & 6.02 & 11.09 \\ 
    InternVL3.5-8B~\cite{wang2025internvl3_5}               & 52.48 & 57.93 & 53.84 & 54.21 & 23.40 & 41.71 & 16.67 & 25.89 & 41.78 \\ 
    InternVL3.5-30B-A3B~\cite{wang2025internvl3_5}           & 59.21 & 59.59 & 58.02 & 58.95 & 22.76 & 45.57 & 16.44 & 26.56 & 44.73 \\ 
    MiniCPM-V4.5-8B~\cite{minicpmv45}                        & 22.31 & 20.62 & 11.95 & 18.82 & 1.72 & 7.29 & 16.78 & 8.17 & 14.14 \\
    \midrule
    \rowcolor{gray!15}
    \multicolumn{10}{l}{\textit{GUI-specific Models (SFT)}} \\
    UI-TARS-7B~\cite{UITARS1}                                & 54.62 & 67.58 & 55.78 & 58.11 & 17.98 & 33.71 & 0.00 & 16.07 & 39.65 \\
    UI-TARS-72B~\cite{UITARS1}                               & 59.65 & \ul{79.14} & \ul{68.32} & 66.96 & \textbf{32.16} & 47.00 & 0.00 & \ul{25.31} & \ul{48.67} \\
    UGround-7B~\cite{uground}                                & 55.06 & 68.06 & 61.03 & 59.99 & 13.37 & 38.86 & 0.00 & 15.51 & 40.46 \\
    \textsc{Jedi}-7B~\cite{Xie2025jediosworld}               & 51.54 & 60.55 & 55.10 & 54.78 & \ul{31.04} & \textbf{54.00} & \textbf{11.00} & \textbf{30.31} & 44.04 \\
    OS-Atlas-7B~\cite{Wu2024OSATLASScreenspotv2}                       & 42.68 & 35.64 & 34.11 & 38.42 & 11.74 & 25.86 & \ul{5.44} & 13.29 & 27.38 \\
    Aguvis-7B~\cite{Xu2024AguvisUP}                         & 34.76 & 36.00 & 34.69 & 35.04 & 11.92 & 30.57 & 0.00 & 12.78 & 25.27 \\
    CogAgent-9B~\cite{hong2024cogagent}                         & 21.06 & 24.67 & 24.49 & 22.95 & 7.68 & 31.00 & 0.00 & 11.16 & 17.77 \\
    OpenCUA-7B~\cite{wang2025opencua}                        & \ul{62.23} & \textbf{84.39} & 67.44 & \ul{69.15} & 21.32 & 49.14 & 0.00 & 21.43 & 48.20 \\
    OpenCUA-32B~\cite{wang2025opencua}                       & \textbf{65.49} & 78.55 & \textbf{68.80} & \textbf{69.64} & 29.09 & \ul{51.00} & 0.00 & 25.08 & \textbf{50.08} \\

    \midrule
    \rowcolor{gray!15}
    \multicolumn{10}{l}{\textit{GUI-specific Models (RL)}} \\
    SE-GUI-7B~\cite{Yuan2025segui}                             & 60.15 & 68.53 & 59.28 & 61.93 & 14.36 & 45.14 & \ul{10.56} & 21.06 & 43.98 \\
    GUI-G$^2$-7B~\cite{tang2025guig2}                        & 68.07 & 80.33 & 71.33 & 72.02 & 23.49 & 51.43 & 0.00 & 22.90 & 50.46 \\
    GTA1-7B~\cite{Yang2025GTA1GT}                            & 63.73 & 76.64 & 57.05 & 64.87 & 23.31 & 51.14 & 0.00 & 22.75 & 46.38 \\
    GTA1-32B~\cite{Yang2025GTA1GT}                           & 75.36 & 88.08 & 76.77 & 78.87 & 38.84 & 67.14 & 0.00 & 33.25 & 58.84 \\
    UI-TARS-1.5-7B~\cite{ui-tars-15-seed}                    & 56.57 & 71.16 & 53.16 & 59.09 & 17.89 & 37.57 & 0.22 & 17.10 & 40.65 \\
    UI-Venus-Ground-7B~\cite{Gu2025UIVenusTR}                & 64.30 & 78.78 & 67.15 & 68.66 & 24.39 & 53.85 & 0.00 & 23.90 & 49.01 \\ 
    UI-Venus-Ground-72B~\cite{Gu2025UIVenusTR}               & \ul{81.58} & \ul{91.30} & \ul{78.81} & \ul{83.12} & \textbf{46.16} & \ul{68.86} & \textbf{51.33} & \textbf{53.75} & \textbf{70.23} \\ 
    Holo1.5-7B~\cite{hai2025holo15modelfamily}               & 62.98 & 80.21 & 74.34 & 70.54 & 28.91 & 55.00 & 0.00 & 26.04 & 51.00 \\
    Holo1.5-72B~\cite{hai2025holo15modelfamily}              & \textbf{81.77} & \textbf{92.01} & \textbf{84.84} & \textbf{85.17} & \ul{40.38} & \textbf{69.43} & 0.00 & \ul{34.47} & \ul{62.91} \\

    \bottomrule
    \end{tabular*}
    \vspace{0.4em}
    \caption{Performance comparison on \textbf{VenusBench-GD} dataset categorized by the evaluation tasks. In the above table, within each group, the best-performing model on each task is highlighted in \textbf{bold}, while the second-best is \ul{underlined}.}
    \label{tab:main_results}
\end{table*}

As shown in Table~\ref{tab:main_results}, the results reveals insights that are difficult to obtain from existing benchmarks:

\vspace{-1mm}
\begin{remark}
General open-source VLMs now match or even surpass specialized GUI models on basic grounding tasks.
\end{remark}
\vspace{-1mm}

Due to the incorporation of extensive GUI-related data during pre-training and instruction tuning, state-of-the-art general vision-language models, particularly the Qwen3-VL series, consistently achieve accuracy above 70\% on basic grounding tasks in average accuracy, outperforming most GUI-specialized models trained via supervised fine-tuning (SFT) or reinforcement learning (RL). This suggests that performance on basic grounding tasks is approaching saturation, and that general models, benefiting from broader world knowledge, hold a distinct advantage in interpreting spatial relations, visual appearance, and OCR content. Consequently, basic tasks alone are no longer sufficient to meaningfully differentiate between general and specialized models, underscoring the need for more nuanced evaluation, precisely the motivation behind VenusBench-GD’s inclusion of advanced tasks.



\vspace{-1mm}
\begin{remark}
Advanced grounding tasks enable a more comprehensive and multidimensional assessment of GUI grounding models.
\end{remark}
\vspace{-1mm}

On UI-intensive reasoning tasks, specifically Functional and Reasoning Grounding, GUI specialized models retain a clear edge: Holo1.5-72B and UI-Venus-Ground-72B achieve over 40\% and 68\% accuracy, respectively, significantly outperforming general VLMs. However, on Refusal Grounding task, most specialized models exhibit almost no ability to abstain from responding to invalid queries; only UI-Venus-Ground-72B attains a modest 51.33\% accuracy. This stark contrast highlights a critical limitation: current GUI-specialized models are prone to overfitting, lacking the robustness and generalization needed to handle out-of-distribution or unanswerable instructions.


\vspace{-2mm}
\subsection{Analysis}

\begin{figure}[t]
    \begin{center}
        \includegraphics[width=0.8\linewidth]{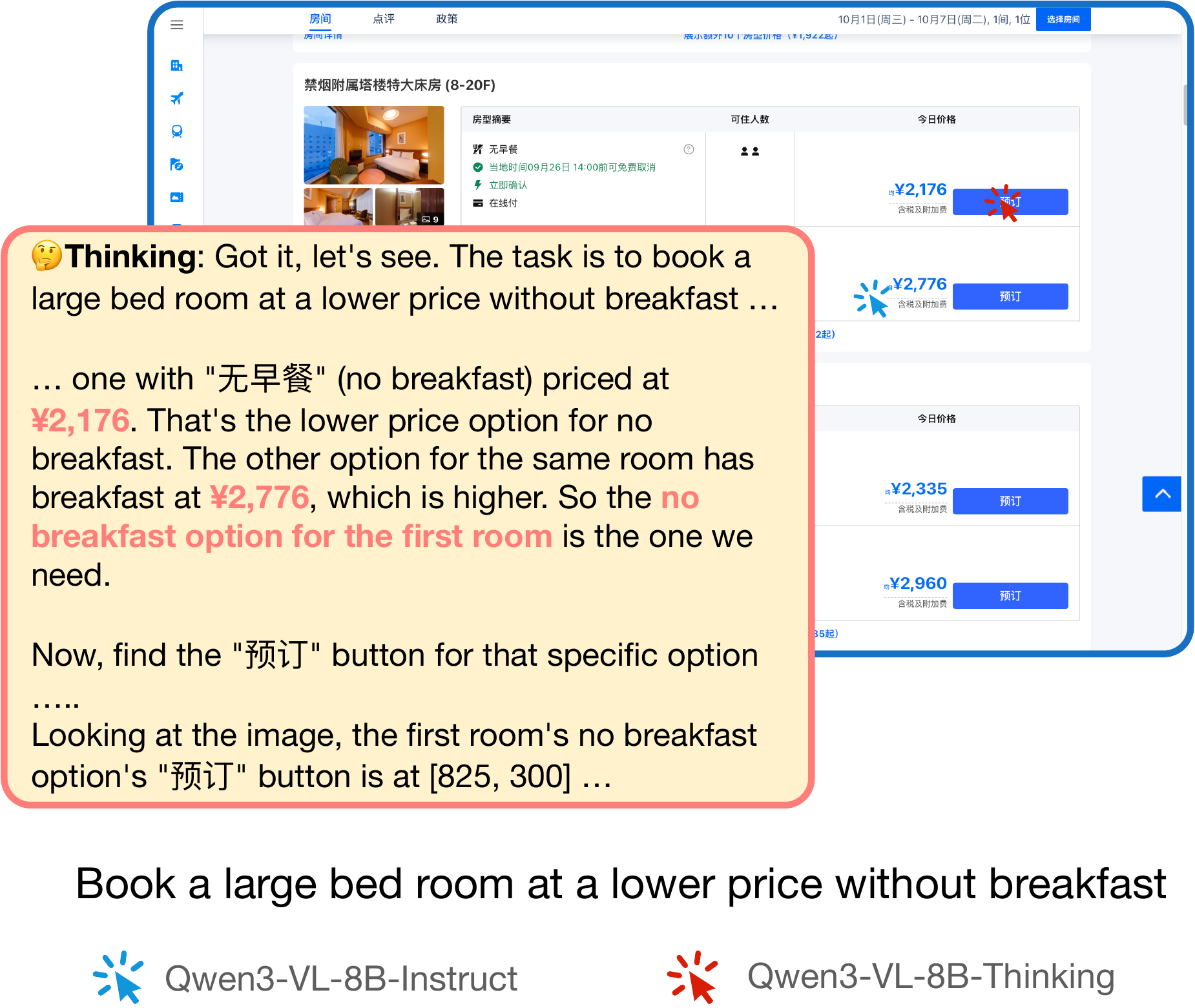}
    \end{center}
	\vspace{-5mm}
    \caption{Thinking-enabled model makes the correct grounding action with detailed analysis of the whole screenshot.}
    \label{fig:reason_vis}
\end{figure}

\noindent \textbf{Model Thinking:} On ScreenSpot-Pro and OSWorld-G, Qwen3-VL-8B-Thinking~\cite{qwen3technicalreport} achieves 46.6\%/56.7\%, underperforming Qwen3-VL-8B-Instruct (54.6\%/58.2\%), a trend consistent with prior findings~\cite{Gu2025UIVenusTR} that explicit reasoning can degrade performance on standard GUI grounding tasks. This motivates our focused comparison between the two variants to investigate whether “thinking” capabilities are better suited for advanced grounding tasks, which demand complex reasoning, rather than basic grounding scenarios. As shown in Figure~\ref{fig:reason_vis}, the thinking-enabled model can effectively analyze the overall interface and use reasoning ability to select the most appropriate click location, indicating that additional reasoning improves performance only in more complex scenarios.

\noindent \textbf{Multilingual Evaluation:} 
A comparative analysis of the model's performance on basic tasks under English and Chinese instruction revealed that the model typically scored higher in the Chinese condition, while its performance on advanced tasks fluctuated, possibly due to instruction ambiguity. As shown in Table~\ref{tab:bilingual_evaluation}, Qwen3-VL-4B demonstrates a notable improvement in basic understanding (from 72.54 to 81.32), suggesting better alignment with Chinese spatial and semantic expressions in simple localization tasks. More detailed results can be found in the supplementary material.

\begin{table}[t]
\centering
\setlength{\tabcolsep}{4pt} 
\renewcommand{\arraystretch}{1.1} 
\resizebox{\columnwidth}{!}{
\begin{tabular}{lccc}
\toprule
\textbf{Models} & \textbf{Basic} & \textbf{Advanced} & \textbf{Overall} \\
\midrule
Qwen3-VL-4B         & 72.54 / 81.32 & 35.20 / 34.94 & 56.15 / 60.95 \\
Holo1.5-7B          & 70.54 / 78.69 & 26.04 / 27.81 & 51.00 / 56.35 \\
UI-Venus-Ground-72B & 83.12 / 86.24 & 53.75 / 51.93 & 70.23 / 71.18 \\
\bottomrule
\end{tabular}%
}
\caption{Performance comparison of representative models in two tasks under English (left) and Chinese (right) instruction.}
\label{tab:bilingual_evaluation}
\vspace{-5mm}
\end{table}

\section{Conclusion}

In this paper, we introduced VenusBench-GD, a large-scale, high-fidelity benchmark spanning web, mobile, and desktop platforms with 97 real-world applications across 10 domains. By incorporating 6 hierarchically designed grounding tasks (from basic localization to complex reasoning/refusal scenarios), VenusBench-GD establishes a rigorous multi-dimensional evaluation framework. 
Combined with a human-AI collaborative annotation pipeline ensuring minimal label noise, our benchmark achieves superior data quality and cross-lingual relevance, which enables meaningful differentiation and valid evaluation between state-of-the-art GUI agents.
\newpage
{
    \small
    \bibliographystyle{ieeenat_fullname}
    \bibliography{main}
}

\clearpage
\setcounter{page}{1}
\maketitlesupplementary

\section{Data Privacy}

All screenshots used in this benchmark were collected from freshly created, synthetic user accounts specifically set up for data annotation purposes. Any user-facing contents, such as names, messages, or profile details, were manually fabricated and do not correspond to any real individual or entity. Consequently, VenusBench-GD contains no personally identifiable information and poses no privacy or compliance risks.

Furthermore, all application interfaces were accessed and captured in accordance with the terms of service of the respective platforms for non-commercial research use. This controlled data generation protocol ensures that our benchmark adheres to ethical standards for responsible AI dataset publication.

\section{Dataset Statistics}

\noindent \textbf{Application Distribution:} We collect applications from three distinct platforms, spanning 10 diverse domains: Creative, Deployment, E-commerce, Entertainment, Finance, Knowledge, Productivity, Social, Travel, and Utility. This broad coverage ensures a rich and representative application distribution, as detailed in Table~\ref{tab:domains_applications_with_superscripts}.

{
\setcellgapes{1.5pt}
\makegapedcells

\begin{table}[t]
\centering
\footnotesize
\setlength{\tabcolsep}{3pt}
\begin{tabular}{c c l}
\toprule
\textbf{Domain} & \textbf{Applications} \\
\midrule

Creative & \makecell[l]{AE$^{3}$, Canva$^{3}$, Figma$^{3}$, Illustrator$^{3}$, \\ Pixso$^{3}$, Sketch$^{3}$} \\
Development & \makecell[l]{Anaconda$^{3}$, CLion$^{3}$, DockerMac$^{3}$, Eclipse$^{3}$, \\ Github$^{1}$, GithubDesktop$^{3}$, IntelliJIDEA$^{3}$, \\ Pycharm$^{3}$, Vscode$^{3}$} \\
E-commerce & \makecell[l]{Amazon$^{1}$, Eaby$^{1}$, Ebay$^{1,2}$, Eleme$^{2}$, Jd$^{1,2}$, \\ Meituan$^{2}$, Pdd$^{2}$, Taobao$^{1,2}$, Yelp$^{2}$} \\
Entertainment & \makecell[l]{NeteaseMusic$^{1,2}$, PodcastPlayer$^{2}$, Spotify$^{1,2}$, \\ Steam$^{1}$, Youtube$^{1,2}$} \\
Finance & \makecell[l]{Alipay$^{2}$, Bloomberg$^{1}$, GoogleWallet$^{2}$, \\ TradingView$^{1}$, WechatPay$^{2}$, YahooFinance$^{1,2}$} \\
Knowledge & \makecell[l]{Obsidian$^{3}$, Qidian$^{2}$, Wikipedia$^{1,2}$} \\
Productivity & \makecell[l]{AsanaCh$^{1}$, AsanaEn$^{1}$, Coze$^{1}$, Deepseek$^{1}$, Evernote$^{3}$, \\ Excel$^{2}$, Jira$^{1}$, Notion$^{3}$, Openai$^{1}$, PowerPoint$^{3}$, \\ Rednote$^{1,2}$, Trello$^{1}$, Typora$^{3}$, Word$^{3}$, Yuque$^{3}$} \\
Social & \makecell[l]{Bilibili$^{1,2}$, Douyin$^{2}$, HackerNews$^{1}$, Instagram$^{2}$, \\ QQNews$^{2}$, Reddit$^{1,2}$, Toutiao$^{2}$, Wechat$^{2}$, \\ Weibo$^{1,2}$, Zhihu$^{1}$} \\
Travel & \makecell[l]{Airbnb$^{1,2}$, Amap$^{1,2}$, Booking$^{1}$, Ctrip$^{1,2}$, Didi$^{2}$, \\ Fliggy$^{2}$, GoogleMaps$^{2}$, GoogleMapsCh$^{1}$, \\ GoogleMapsEn$^{1}$, Skyscanner$^{1}$, Uber$^{2}$} \\
Utility & \makecell[l]{Alarm$^{2}$, SystemEn$^{2}$, SystemZh$^{2}$, TodoList$^{2}$, \\ Weather$^{2}$, Windy$^{2}$} \\
\bottomrule
\end{tabular}
\caption{A list of all domains defined in VenusBench-GD, along with the corresponding applications and their associated platforms. Superscripts 1, 2, and 3 denote applications on web, mobile, and desktop platforms, respectively.}
\label{tab:domains_applications_with_superscripts}
\vspace{-3mm}
\end{table}
}

\begin{figure*}[!htbp]
    \begin{center}
        \includegraphics[width=1.0\linewidth]{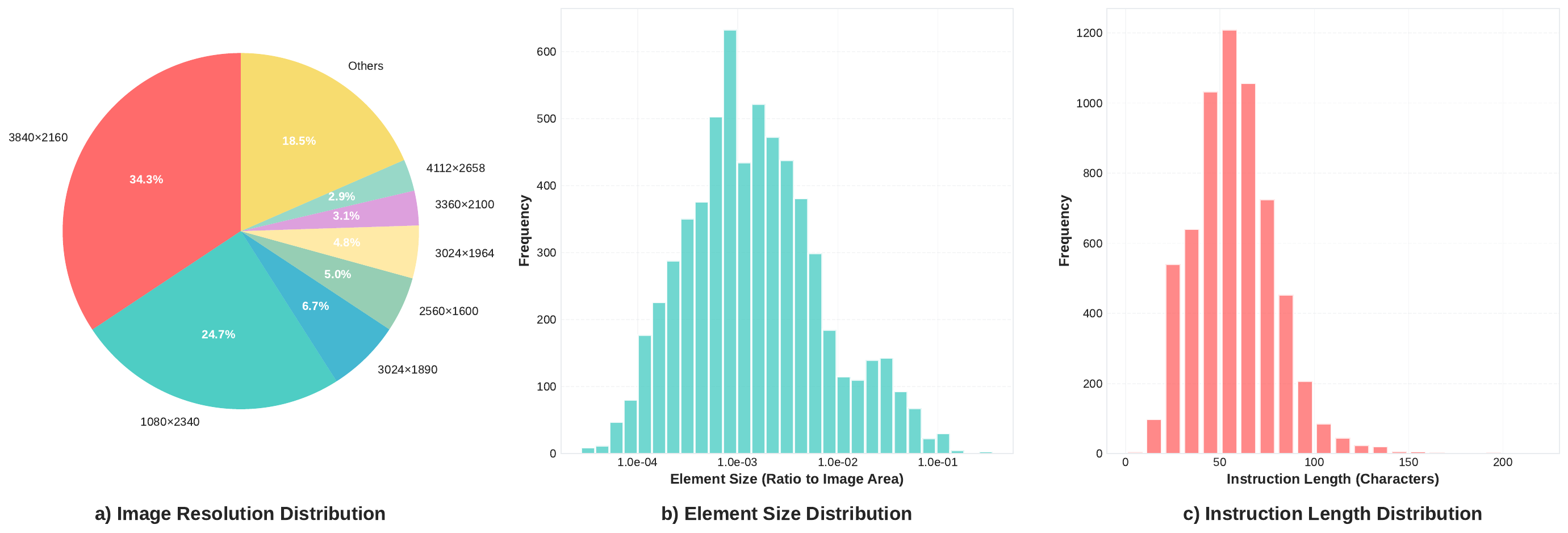}
    \end{center}
	\vspace{-5mm}
    \caption{\textbf{The dataset statistics of VenusBench-GD.} a) Distribution of image resolution sizes. For clarity in visualization, we report only the top 7 most frequent image resolutions observed in the benchmark; all remaining resolutions are aggregated into the "Others" category for statistical completeness. b) Distribution of the element size relative to image size. c) Distribution of the instruction length.}
    \label{fig:benchmark_statistics}
\end{figure*}

\noindent \textbf{Benchmark Statistics:} We show the distributions of image resolutions, UI element sizes, and instruction lengths in VenusBench-GD as shown in Figure~\ref{fig:benchmark_statistics}. By design, VenusBench-GD ensures high diversity across these dimensions, reflecting its commitment to comprehensive and representative benchmarking of GUI grounding capabilities.

\noindent \textbf{Additional Benchmark Comparisons:} We additionally include the MMBench-GUI~\cite{Wang2025MMBenchGUIHM} benchmark and perform a statistical analysis of its annotation quality to evaluate labeling consistency and reliability as shown in Table~\ref{tab:benchmark_comparison_add}.

\begin{table*}[t]
\centering
\footnotesize
\setlength{\tabcolsep}{3pt}
\begin{tabular*}{\textwidth}{@{\extracolsep{\fill}}l *{9}{c}}
\toprule
 & \multicolumn{2}{c}{\textbf{Environments}} & \multicolumn{4}{c}{\textbf{Quality Analysis$\downarrow$}} & \multicolumn{3}{c}{\textbf{Statistics$\uparrow$}} \\
\cmidrule(lr){2-3} \cmidrule(lr){4-7} \cmidrule(lr){8-10}
\textbf{Benchmarks} & \textbf{Platform} & \textbf{Language} &
\textbf{IL} & \textbf{AR} & \textbf{SM} &\textbf{Overall} &
\textbf{\# Sample} & \textbf{Avg. Ele} & \textbf{Apps} \\
\midrule
ScreenSpot-V2~\cite{Wu2024OSATLASScreenspotv2} & Desktop, Web, Mobile & EN & 9.2\% & 0.8\% & 0.8\% &10.8\% & 1,272 & 1.74 & N/A \\
ScreenSpot-Pro~\cite{Li2025ScreenSpotProGG} & Desktop & EN,CN & 2.2\% & 1.8\% & 0.6\% &4.6\% & 1,581 & 1.0 & 23 \\
OSWorld-G~\cite{Xie2025jediosworld} & Desktop & EN & 9.2\% & 2.6\% & 2.4\% &14.2\% & 564 & 2.25 & 8 \\
UI-Vision~\cite{Nayak2025UIVisionAD} & Desktop & EN & 6.0\% & 2.4\% & 2.0\% &10.4\% & 5,479* & \textbf{4.64} & 83 \\
CAGUI~\cite{Zhang2025AgentCPMGUIBM} & Mobile & CN & 11.6\% & 6.4\% & 6.6\% &24.6\% & 3,000 & 1.0 & N/A \\
MMBench-GUI~\cite{Wang2025MMBenchGUIHM} & Desktop, Web, Mobile & EN & 3.2\% & 3.0\% & 2.6\% &8.8\% & 3574 & 2.77 & N/A \\
\midrule
VenusBench-GD (ours) & Desktop, Web, Mobile & EN, CN & \textbf{1.0\%} & \textbf{0.8\%} & \textbf{0.8\%} &\textbf{2.6\%} & \textbf{6,166} & 2.83 & \textbf{97} \\
\bottomrule
\end{tabular*}
\vspace{-0.2em}
\caption{Comparison of existing GUI grounding benchmarks.
}
\label{tab:benchmark_comparison_add}
\end{table*}

\section{Benchmark Tasks}

\noindent \textbf{Dense Elements and Accurate Localization:} We present the \textbf{13} distinct UI element types annotated in VenusBench-GD shown in Table~\ref{tab:ui_element_types}, along with representative visual examples. To reduce annotation complexity, we merged semantically similar and commonly co-occurring elements, such as notification and dialog, or checkbox and radiobox, into unified categories.

Our benchmark construction follows a dense annotation protocol: each screenshot is exhaustively labeled with approximately \textbf{20} UI elements, ensuring broad coverage of UI categories across the dataset. Annotators were instructed to delineate bounding boxes that closely encompass the full interactive region of each element, prioritizing tight alignment and completeness.

The annotation process underwent three iterative rounds of refinement and quality control. Inter-round validation yielded steadily improving accuracy rates of \textbf{84.54\%}, \textbf{95.47\%}, and \textbf{98.25\%}, respectively, demonstrating the high fidelity and reliability of the final annotations. The validation accuracy was determined by independent reviewers who manually inspected a randomly sampled subset of approximately 15\% of the annotated data.

{
\setcellgapes{3pt} 
\makegapedcells

\begin{table}[t]
\centering
\footnotesize
\setlength{\tabcolsep}{5pt} 
\begin{tabular}{l c l c}
\toprule
\textbf{Element Type} & \textbf{Visualization} & \textbf{Element Type} & \textbf{Visualization} \\
\midrule

icon & \includegraphics[width=0.08\linewidth,valign=c]{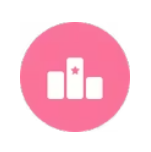} &
dropdown & \includegraphics[width=0.08\linewidth,valign=c]{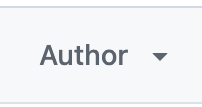} \\

button & \includegraphics[width=0.08\linewidth,valign=c]{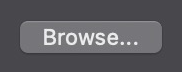} &
image & \includegraphics[width=0.08\linewidth,valign=c]{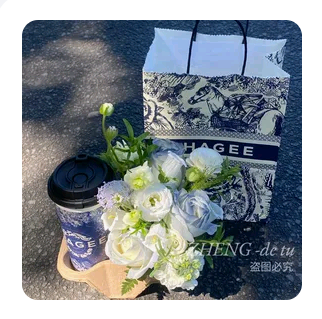} \\

text/tag & \includegraphics[width=0.08\linewidth,valign=c]{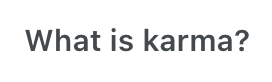} &
text input & \includegraphics[width=0.08\linewidth,valign=c]{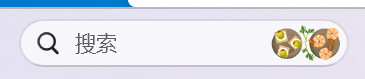} \\

menu & \includegraphics[width=0.08\linewidth,valign=c]{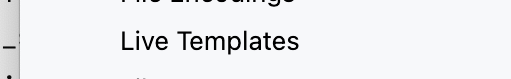} &
link & \includegraphics[width=0.08\linewidth,valign=c]{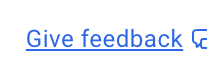} \\

slider & \includegraphics[width=0.08\linewidth,valign=c]{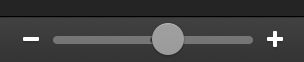} &
notification/dialog & \includegraphics[width=0.08\linewidth,valign=c]{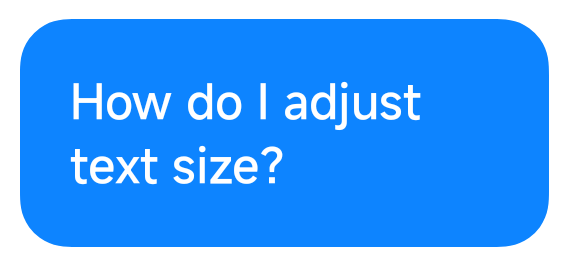} \\

checkbox/radiobox & \includegraphics[width=0.08\linewidth,valign=c]{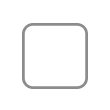} &
toggle switch & \includegraphics[width=0.08\linewidth,valign=c]{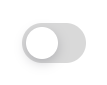} \\

tab & \includegraphics[width=0.08\linewidth,valign=c]{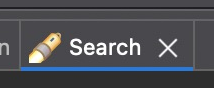} & & \\

\bottomrule
\end{tabular}
\caption{The 13 UI element types used in our dataset, each accompanied by a representative visualization. Two elements are shown per row for compact presentation with improved spacing and alignment. For visualization, we selected a subset of annotated UI elements and expanded their bounding boxes by 20 pixels outward to enhance visibility.}
\label{tab:ui_element_types}
\vspace{-3mm}
\end{table}
}

\begin{figure*}[!htbp]
    \begin{center}
        \includegraphics[width=1.0\linewidth]{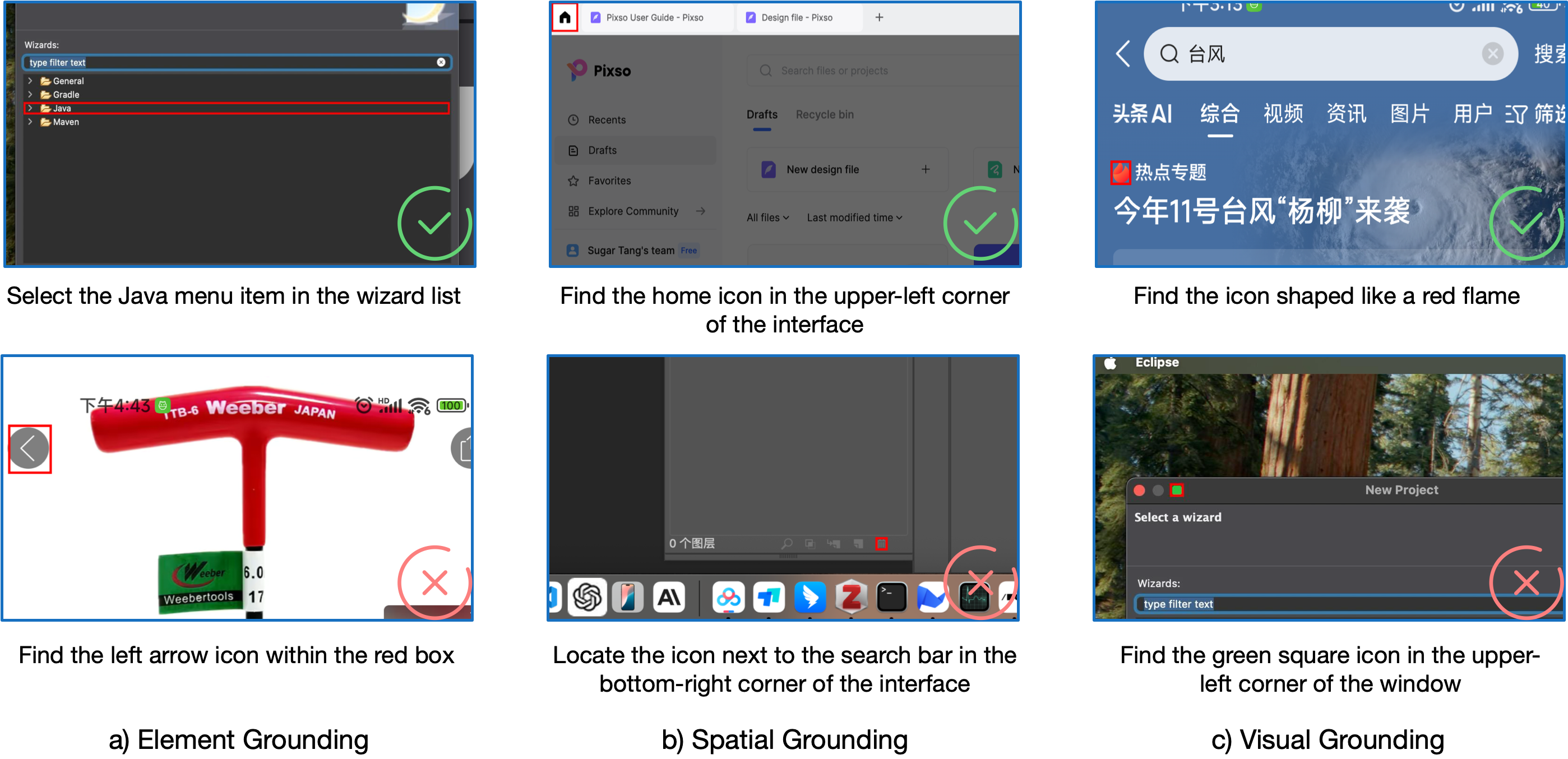}
    \end{center}
	\vspace{-5mm}
    \caption{\textbf{Examples of element grounding tasks,} illustrating both correct and incorrect matches between generated instructions and their corresponding annotated bounding boxes}
    \vspace{-2mm}
    \label{fig:element_examples}
\end{figure*}

\noindent \textbf{Basic Grounding Tasks:} For basic grounding tasks, we generate instructions by leveraging the element category, textual content, visual appearance, and relative spatial context of the target UI element. Specifically, we employ the Qwen2.5-VL-72B model, guided by a carefully designed prompt that explicitly specifies our generation requirements and includes diverse exemplars covering various instruction types. The input images are augmented with a red rectangular bounding box highlighting the target element. After generation, each instruction–UI element pair is manually validated by human annotators for semantic and referential correctness. Furthermore, all generated instructions are categorized into three grounding dimensions: element grounding, spatial grounding, and visual grounding, enabling fine-grained analysis of model capabilities across distinct grounding modalities. We show examples of element grounding tasks in Figure~\ref{fig:element_examples}.

Advanced grounding tasks exhibit greater complexity than basic grounding tasks in both task difficulty and structural variation. To accommodate the diverse cognitive and perceptual demands of these tasks, we employ distinct construction pipelines tailored to each task type:

\noindent \textbf{Reasoning Grounding:}  We pre-selected a set of domains particularly well-suited for reasoning-oriented grounding tasks, namely \textbf{E-commerce, Travel, Finance, Productivity, and Social}, and developed domain-specific instruction templates that encapsulate the characteristic user intents of each scenario as shown in Table~\ref{tab:reason_templates}. Annotators were required to select representative screenshots from these domains and adapt the provided templates to formulate natural, contextually grounded instructions, along with precise bounding box annotations for the target UI elements. This template-driven annotation protocol ensures that the resulting reasoning tasks exhibit both sufficient cognitive complexity and high referential accuracy.

\noindent \textbf{Refusal Grounding:}  For refusal grounding, we deliberately modify existing instruction–UI element pairs by altering key attributes in the instruction, such as spatial references, visual styles, textual content, or the specified UI element type. The revised instruction no longer corresponds to any valid element in the given screenshot and this controlled perturbation ensures that the resulting refusal instructions are semantically plausible yet ungroundable, thereby introducing sufficient ambiguity and perceptual confusion.

\noindent \textbf{Functional Grounding:} To generate accurate instruction, we provided both pre-click and after-click screenshots to Qwen2.5-VL-72B to describe the user intent and the functionality of the specified element. Additionally, to enhance coverage of functionalities in specialized software, annotators were instructed to follow official documentation when crafting or refining grounding instructions.

\begin{figure*}[!htbp]
    \begin{center}
        \includegraphics[width=1.0\linewidth]{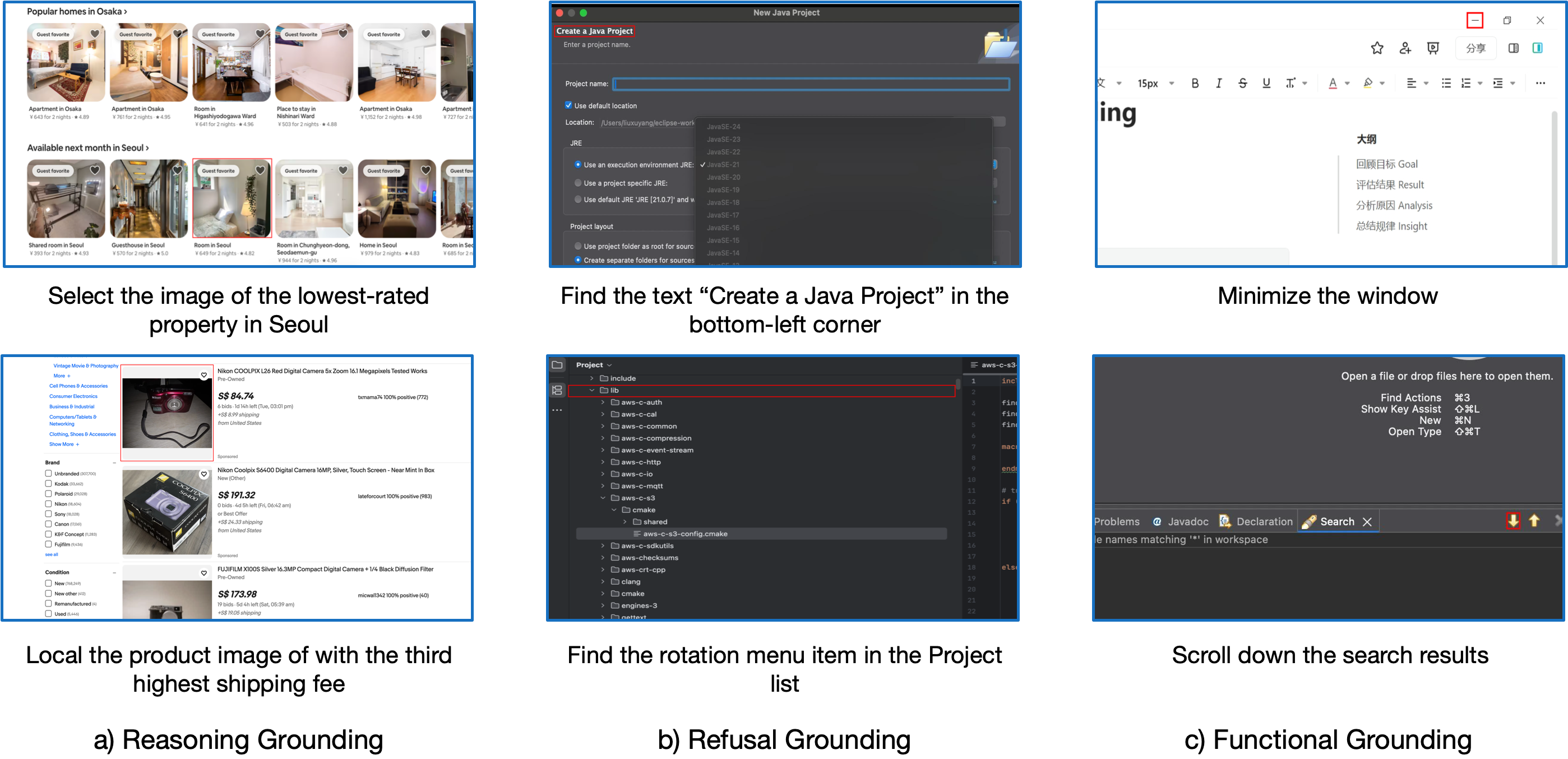}
    \end{center}
	\vspace{-5mm}
    \caption{\textbf{Examples of advanced grounding tasks.} In the refusal grounding task, the red bounding box indicates the original UI element. After modification of the instruction, no matching element exists in the image.}
    \vspace{-2mm}
    \label{fig:advanced_examples}
\end{figure*}

We show examples of advanced grounding tasks in Figure~\ref{fig:advanced_examples}.

{
\setcellgapes{1.5pt}
\makegapedcells

\begin{table}[t]
\centering
\footnotesize
\setlength{\tabcolsep}{4pt}
\begin{tabularx}{\linewidth}{@{} c X @{}}
\toprule
\textbf{Domain} & \textbf{Instruction Templates} \\
\midrule

E-commerce & 
1.Find the item with the highest discount amount and add it to the cart. \\
& 2.Find the item with the third-highest number of reviews and add it to the cart. \\
& 3.Select the item with the lowest rating to view its details. \\
\midrule

Finance & 
1.Among all stocks priced below \$50, find the one with the highest market capitalization and view its details. \\
& 2.Identify the U.S. stock with the highest price increase today. \\
& 3.Find the stock with the highest average trading volume over the past three months. \\
\midrule

Productivity & 
1.Locate the task with a deadline of October 15. \\
& 2.Complete the last subtask of ``xxx''. \\
& 3.View the task that took the shortest time to complete in the first half of this year. \\
\midrule

Social & 
1.Find the most recently posted comment below the article. \\
& 2.View the details of the only comment that has been edited. \\
& 3.Find the comment with the highest number of likes. \\
\midrule

Travel & 
1.Find the only flight with exactly two stopovers. \\
& 2.Among all Cathay Pacific flights, identify the one with the longest total flight duration. \\
& 3.Find the most expensive hotel on the map. \\
\bottomrule
\end{tabularx}
\caption{A list of selected domains and examples of corresponding instruction templates.}
\label{tab:reason_templates}
\vspace{-5mm}
\end{table}
}

\noindent \textbf{Post-Selection:} We first gathered predictions from a suite of leading models (Qwen3-VL-8B, UI-TARS-1.5-7B, Holo1.5-7B, GTA1-7B and UI-Venus-Ground-7B) and used their level of agreement as a filtering metric. This process involved pruning most "easy" instances that all models solved correctly, while also flagging instances where models consistently converged as the wrong answer. These latter cases were reviewed by annotators to correct labeling errors to improving the dataset's reliability.

After correction, we classified instances into three difficulty levels based on the proportion of models that predicted correctly—difficult ($\leq20\%$), medium (40–60\%), and easy (all others). Finally, we constructed the benchmark by randomly sampling from these categories in a 4:4:2 ratio.

\section{Experimental Analsysis}

\begin{figure}[t]
    \begin{center}
        \includegraphics[width=1.0\linewidth]{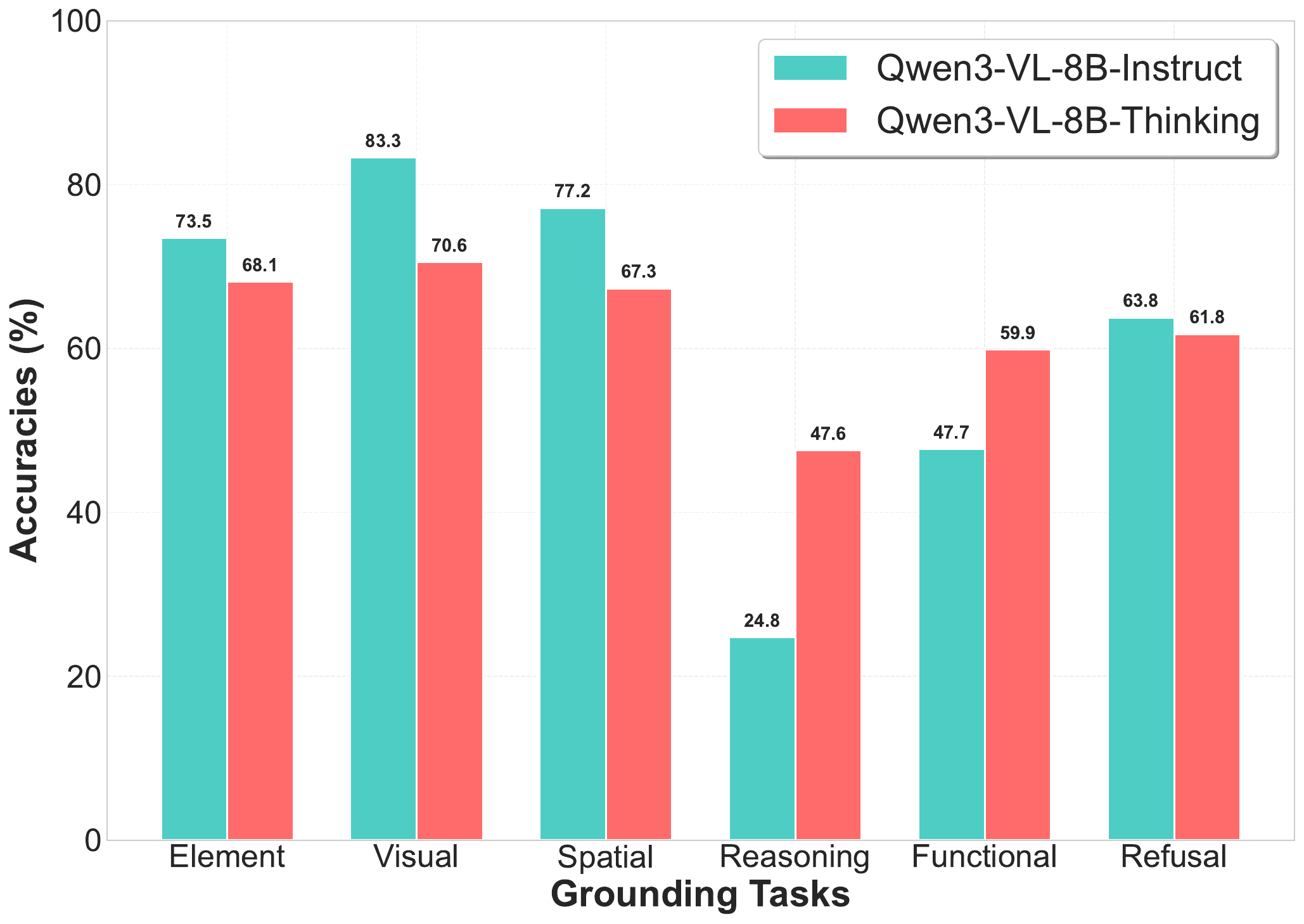}
    \end{center}
	\vspace{-5mm}
    \caption{The model performance comparison of Qwen3-VL-8B-Instruct and Qwen3-VL-8B-Thinking. The reasoning-enhanced thinking model outperforms instruct editions on advanced grounding tasks but lags behind on basic grounding tasks.}
    \label{fig:instruct_vs_thinking}
\end{figure}

\noindent \textbf{Model Thinking:} According to the technical report~\cite{qwen3technicalreport}, Qwen3-VL-8B-Instruct and Qwen3-VL-8B-Thinking achieve grounding scores of \textbf{94.4\%/54.6\%/58.2\%} and \textbf{93.6\%/46.6\%/56.7\%} on ScreenSpot, ScreenSpot-Pro, and OSWorld-G, respectively. Surprisingly, the model with stronger reasoning capabilities exhibits weaker grounding performance, prompting a detailed comparison on VenusBench-GD shown in Figure~\ref{fig:instruct_vs_thinking}.

On basic grounding tasks, the thinking-enabled model exhibits slightly lower performance than the instruct model, a trend consistent with observations from prior benchmarks. In contrast, it significantly outperforms the instruct model on advanced grounding tasks, with notable improvements of \textbf{22.8\% }in reasoning grounding and \textbf{12.2\%} in functional grounding, where successful execution requires multi-step inference and deliberate reasoning. This aligns well with our design intent for advanced tasks. Performance on refusal grounding is comparable between the two models.

These findings suggest that the thinking model is better suited for complex tasks involving extended reasoning and logical deliberation, while its performance on simpler tasks may be adversely affected by redundant or irrelevant outputs and sensitivity to bounding box annotation quality.

\noindent \textbf{Bilingual Evaluation:} We also conduct comparisons on VenusBench-GD using Chinese instructions as shown in Table~\ref{tab:main_results_zh}. Notably, some models exhibit language-dependent performance: for instance, OpenCUA-7B outperforms UI-TARS-7B on English instructions, while the reverse holds for Chinese. This highlights the value of multilingual data in grounding model training.

\begin{table*}[ht]
    \centering
    \footnotesize
    \setlength{\tabcolsep}{0pt}
    \begin{tabular*}{\textwidth}{@{\extracolsep{\fill}}l *{9}{c}}
    \toprule
    \multirow{2}{*}{\textbf{Models}} & 
    \multicolumn{4}{c}{\textbf{Basic Tasks}} & 
    \multicolumn{4}{c}{\textbf{Advanced Tasks}} & 
    \multirow{2}{*}{\textbf{Overall}} \\
    \cmidrule(lr){2-5} \cmidrule(lr){6-9}
                    & \textbf{Element} & \textbf{Visual} & \textbf{Spatial} & \textbf{Avg} &
                      \textbf{Reasoning} & \textbf{Functional} & \textbf{Refusal} & \textbf{Avg} & \\
    \midrule
    \rowcolor{gray!15}
    \multicolumn{10}{l}{\textit{Closed-source Models}} \\
    GPT-4o~\cite{Hurst2024GPT4oSC}                           & 4.71 & 10.73 & 6.61 & 6.73 & 3.79 & 3.71 & \ul{33.00} & \ul{13.48} & 9.69 \\ 
    Gemini 2.5 Pro~\cite{gemini25pro}                        & \ul{6.79} & \ul{13.35} & \ul{12.54} & \ul{10.09} & \ul{7.68} & \ul{9.43} & \textbf{35.55} & \textbf{17.39} & \ul{13.29} \\
    Claude-Sonnet-4.0~\cite{claude4systemcard}               & \textbf{20.05} & \textbf{43.62} & \textbf{19.05} & \textbf{25.46} & \textbf{9.93} & \textbf{12.28} & 18.66 & 13.44 & \textbf{20.18} \\
    \midrule
    \rowcolor{gray!15}
    \multicolumn{10}{l}{\textit{General Open-source VLMs}} \\
    Qwen2.5-VL-7B~\cite{Qwen2.5-VL}                          & 67.76 & 67.70 & 57.34 & 64.64 & 15.26 & 38.57 & 13.77 & 20.79 & 45.39 \\ 
    Qwen2.5-VL-72B~\cite{Qwen2.5-VL}                          & 80.58 & \ul{88.20} & 75.41 & 80.89 & \textbf{35.05} & 52.14 & \textbf{67.44} & \textbf{50.24} & \textbf{67.43} \\ 
    Qwen3-VL-4B~\cite{qwen3technicalreport}                          & \ul{84.66} & 82.96 & 74.83 & 81.32 & 25.56 & \ul{62.57} & 25.00 & 34.94 & 60.95 \\ 
    Qwen3-VL-8B~\cite{qwen3technicalreport}                  & \textbf{85.36} & \textbf{88.92} & \textbf{80.08} & \textbf{84.65} & 22.04 & 50.43 & \ul{55.00} & \ul{40.34} & \ul{65.19} \\ 
    Qwen3-VL-30B-A3B~\cite{qwen3technicalreport}             & 82.84 & 86.89 & \ul{78.52} & \ul{82.54} & \ul{32.16} & \textbf{67.14} & 25.11 & 38.86 & 63.36 \\ 
    InternVL2.5-8B~\cite{InternVL25}                        & 15.78 & 18.36 & 13.22 & 15.64 & 7.40 & 13.43 & 0.00 & 6.50 & 11.63 \\ 
    InternVL3.5-8B~\cite{wang2025internvl3_5}               & 69.26 & 56.50 & 57.24 & 62.59 & 19.60 & 48.00 & 5.11 & 22.12 & 44.82 \\ 
    InternVL3.5-30B-A3B~\cite{wang2025internvl3_5}           & 73.66 & 61.50 & 58.11 & 66.09 & 21.22 & 49.86 & 7.00 & 23.89 & 47.56 \\ 
    MiniCPM-V4.5-8B~\cite{minicpmv45}                        & 38.59 & 27.77 & 20.51 & 30.59 & 5.23 & 18.00 & 8.33 & 9.56 & 21.35 \\
    \midrule
    \rowcolor{gray!15}
    \multicolumn{10}{l}{\textit{GUI-specific Models (SFT)}} \\
    UI-TARS-7B~\cite{UITARS1}                                & 68.01 & 72.47 & 61.71 & 67.22 & 16.17 & 37.71 & 0.22 & 16.43 & 44.91 \\
    UI-TARS-72B~\cite{UITARS1}                               & \textbf{72.72} & \textbf{84.03} & \textbf{73.28} & \textbf{75.63} & \textbf{34.96} & \textbf{59.42} & 0.00 & \textbf{29.66} & \textbf{55.44} \\
    UGround-7B~\cite{uground}                                & 65.81 & 69.01 & 59.86 & 64.82 & 16.17 & 48.43 & 0.00 & 19.13 & 44.75 \\
    \textsc{Jedi}-7B~\cite{Xie2025jediosworld}               & 45.82 & 52.56 & 47.33 & 47.90 & 23.21 & 54.71 & \textbf{6.55} & \ul{25.82} & 38.21 \\
    OS-Atlas-7B~\cite{Wu2024OSATLASScreenspotv2}                       & 50.72 & 33.85 & 33.82 & 41.60 & 12.01 & 28.43 & \ul{5.44} & 14.07 & 29.51 \\
    Aguvis-7B~\cite{Xu2024AguvisUP}                         & 43.24 & 35.88 & 37.71 & 39.81 & 11.11 & 33.00 & 0.00 & 13.07 & 28.07 \\
    CogAgent-9B~\cite{hong2024cogagent}                         & 37.08 & 25.27 & 35.18 & 33.65 & 5.69 & 29.86 & 0.00 & 10.05 & 23.29 \\
    OpenCUA-7B~\cite{wang2025opencua}                        & \ul{71.15} & \ul{81.88} & \ul{66.57} & \ul{72.39} & 22.13 & \ul{57.28} & 0.00 & 23.86 & \ul{51.08} \\
    OpenCUA-32B~\cite{wang2025opencua}                       & 59.46 & 68.65 & 59.48 & 61.69 & \ul{28.18} & 48.14 & 0.33 & 24.08 & 45.18 \\

    \midrule
    \rowcolor{gray!15}
    \multicolumn{10}{l}{\textit{GUI-specific Models (RL)}} \\
    SE-GUI-7B~\cite{Yuan2025segui}                             & 71.15 & 73.30 & 61.32 & 68.75 & 15.08 & 50.28 & \ul{7.44} & 21.64 & 48.07 \\
    GUI-G$^2$-7B~\cite{tang2025guig2}                        & 79.45 & 83.08 & 74.54 & 78.87 & 25.74 & 58.14 & 0.00 & 25.56 & 55.46 \\
    GTA1-7B~\cite{Yang2025GTA1GT}                            & 73.29 & 77.47 & 59.48 & 70.19 & 20.42 & 55.71 & 0.00 & 22.75 & 49.36 \\
    GTA1-32B~\cite{Yang2025GTA1GT}                           & 83.09 & 90.46 & 78.23 & 83.43 & 36.67 & 68.85 & 0.00 & 32.80 & 61.20 \\
    UI-TARS-1.5-7B~\cite{ui-tars-15-seed}                    & 72.16 & 71.28 & 58.41 & 67.85 & 18.25 & 44.14 & 0.11 & 18.91 & 46.36 \\
    UI-Venus-Ground-7B~\cite{Gu2025UIVenusTR}                & 74.73 & 78.31 & 68.03 & 73.61 & 24.12 & 58.57 & 0.00 & 25.00 & 52.54 \\ 
    UI-Venus-Ground-72B~\cite{Gu2025UIVenusTR}               & \ul{87.18} & \textbf{91.42} & \ul{80.56} & \ul{86.24} & \textbf{46.43} & \ul{71.71} & \textbf{43.33} & \textbf{51.93} & \textbf{71.18} \\ 
    Holo1.5-7B~\cite{hai2025holo15modelfamily}               & 79.07 & 81.88 & 75.51 & 78.69 & 26.74 & 65.28 & 0.00 & 27.81 & 56.35 \\
    Holo1.5-72B~\cite{hai2025holo15modelfamily}              & \textbf{89.19} & \ul{91.30} & \textbf{86.88} & \textbf{89.01} & \ul{38.93} & \textbf{75.28} & 0.00 & \ul{35.38} & \ul{65.46} \\

    \bottomrule
    \end{tabular*}
    \vspace{0.4em}
    \caption{Performance comparison of \textbf{Chinese} instructions on \textbf{VenusBench-GD} dataset categorized by the evaluation tasks. In the above table, within each group, the best-performing model on each task is highlighted in \textbf{bold}, while the second-best is \ul{underlined}.}
    \label{tab:main_results_zh}
\end{table*}

\begin{figure}[t]
    \begin{center}
        \includegraphics[width=1.0\linewidth]{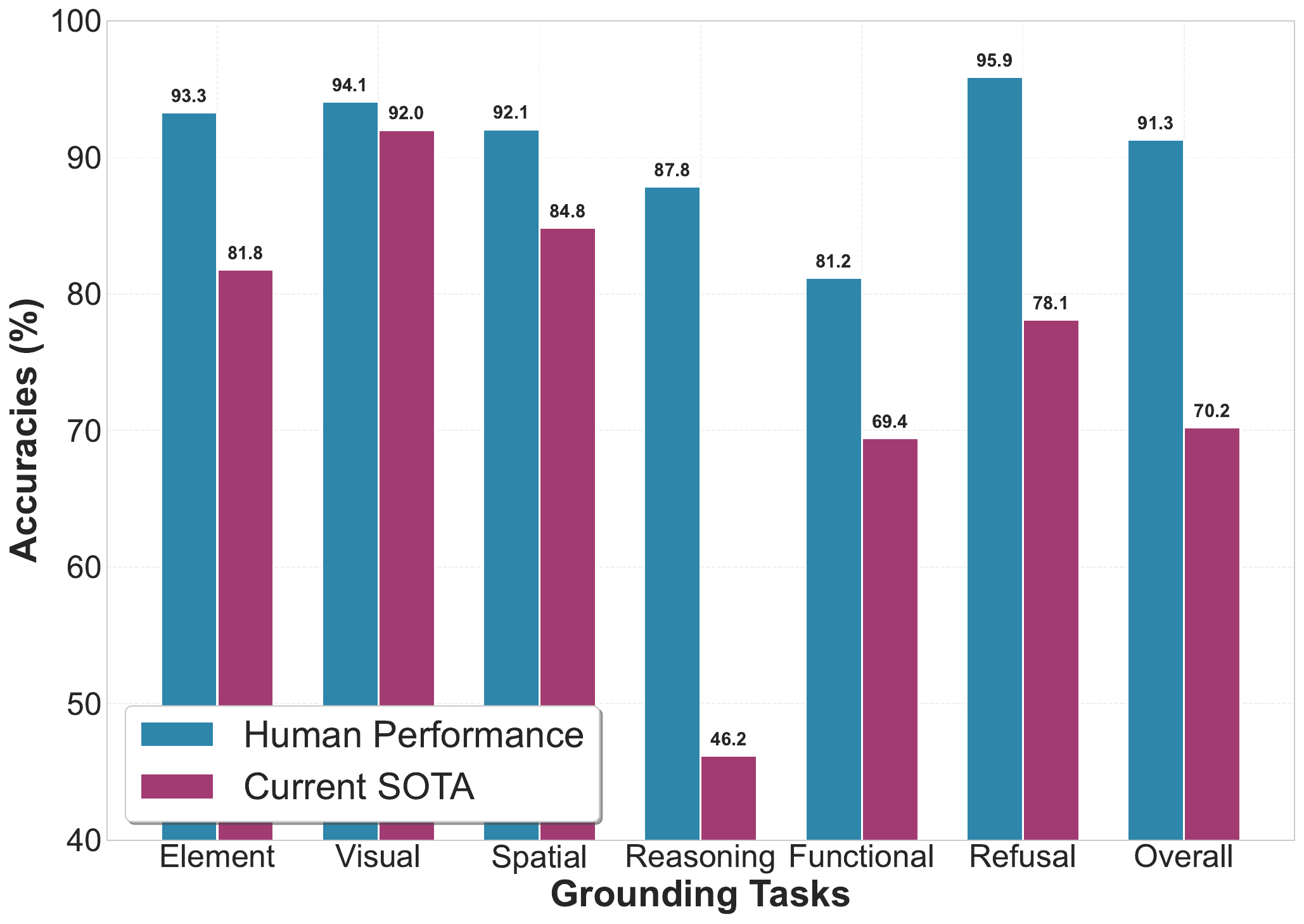}
    \end{center}
	\vspace{-5mm}
    \caption{Human performance comparison with current SOTA performance on grounding tasks.}
    \label{fig:human_performance}
\end{figure}

\noindent \textbf{Human Performance:} We also compare human performance on VenusBench-GD against current SOTA performance shown in see Figure~\ref{fig:human_performance}. Humans significantly outperform all models on advanced tasks, exceeding SOTA by 41.6\%, 11.8\%, and 17.8\% in reasoning, functional, and refusal grounding, respectively. This underscores the limitations of existing models in handling complex, reasoning intensive grounding tasks.

\section{Error Analaysis} 

\textbf{A. Semantic Misinterpretation:} Models struggle with element grounding when an object's identity relies on abstract semantic concepts rather than literal text. For instance, they fail to associate the abstract notion of "text alignment" with its conventional iconography. Failures also arise from OCR limitations on small or stylized text, hindering the bridging of high-level semantics with low-level visual features.

\textbf{B. Imprecise Spatial Localization:} In spatial grounding, models exhibit an inability to integrate spatial constraints with precise object identification, especially in cluttered UI regions. While capable of coarse-grained regional identification (e.g., "upper right corner"), they often fail to distinguish the target element from adjacent distractors, indicating a weakness in jointly optimizing for object identity ("what") and location ("where").

\textbf{C. Failure in Composing Visual Attributes:} For visual grounding, models often fail to compose multiple visual attributes (e.g., color, shape, relation) as specified in an instruction. A common error is latching onto a single salient feature while ignoring others, leading to the selection of partially matching elements. This reveals a fundamental weakness in binding a complete set of visual attributes to a single target.

\textbf{D. Deficiencies in Multi-Step Logical Reasoning:} In reasoning grounding, models demonstrate an inability to execute multi-step logical procedures such as comparison, filtering, or ordinal selection. They often exhibit "greedy" behavior, failing to decompose complex instructions into a sequence of required operations (e.g., selecting the highest-priced item instead of the second-highest).

\textbf{E. Lack of Abstract Functional Knowledge:} During functional grounding, models fail to map abstract, goal-oriented commands (e.g., "Enter Fullscreen") to their non-literal UI affordances. Lacking the domain-specific knowledge of software conventions that humans possess, models resort to futile literal text searches, highlighting a gap between abstract user intent and concrete UI interactions.

\textbf{F. Failure in Contradiction Detection and Task Refusal:} A critical deficiency in refusal grounding is the model's over-compliance with instructions that are factually impossible or contradictory to the visual context.  Given a factually incorrect instruction, such as "Find the text at 2:35 PM in the upper right corner" when the element is actually in the upper left, the model is expected to identify the contradiction and refuse the request. However, a prevalent error is that the model attempts to "hallucinate" a result. It either points to the instructed region (the upper right corner) despite the object's absence or incorrectly identifies a different object that does not match the description. This indicates an over-reliance on the textual prompt and a critical weakness in validating instructions against visual evidence. To mitigate this, models could be trained with contrastive examples (correct vs. incorrect prompts) or by implementing a preliminary verification step to confirm the prompt's premises before execution.

\section{Human Annotation} 

We collaborated with a dedicated annotation team to construct this benchmark dataset. The team comprised 20 members: 17 primary annotators responsible for core labeling tasks, 2 quality assurance specialists handling validation and acceptance, and 1 project coordinator overseeing progress and timelines. All annotators were aged between 25 and 35 and underwent approximately two weeks of intensive training and certification prior to the formal annotation phase.

The entire annotation process spanned roughly three months. We first collected screenshots from diverse platforms and software applications and conducted labeling on a custom built annotation platform. In the initial stage, bounding boxes were drawn around available UI elements, yielding an average of 20 annotated elements per image. Subsequently, distinct annotation pipelines were applied according to the specific grounding task type. For each subtask, we implemented at least two rounds of annotation followed by rigorous quality checks. To further enhance diversity and accuracy, we integrated large language models into the workflow, leveraging their capabilities for instruction generation, refinement, and validation.

\end{document}